\documentclass[twoside,11pt]{article}

\usepackage{array}
 \usepackage{times}
\usepackage{hyperref}       
\hypersetup{
    colorlinks,
    linkcolor=blue, 
    citecolor=blue, 
    linktoc=all 
  }

\usepackage{natbib}

\usepackage{microtype}
\usepackage{graphicx}
\usepackage{subfigure}
\usepackage{booktabs} 

\usepackage{hyperref}
\usepackage{algorithm}
\usepackage{algorithmic}

\usepackage{url}
\usepackage{amssymb,dsfont,amsmath,amsthm,mathtools,mathrsfs}

\usepackage{enumitem}
\usepackage{bbm}
\usepackage{dcolumn}
\usepackage{tikz}
\usetikzlibrary{positioning}
\usepackage{color}

\usepackage{xspace}
\newcolumntype{d}[1]{D{.}{.}{#1}}

\renewcommand{\algorithmicrequire}{\textbf{Input:}}
\renewcommand{\algorithmicensure}{\textbf{Output:}}

\newcommand{\norm}[1]{\|#1\|}
\def\R{\mathbb{R}}
\def\1{\mathbbm{1}}

\def\X{\mathbf{X}}
\def\u{\mathbf{u}}
\def\v{\mathbf{v}}
\def\y{\mathbf{y}}

\def\E{\mathbf{E}}

\def\s{\mathbf{s}}

\def\P{\mathbf{P}}

\def\V{\mathbf{V}}

\newcommand{\Ex}{{\rm I\kern-.3em E}}

\def\cP{\mathcal{P}}

\newcommand{\diff}{\textrm{d}}

\usepackage[colorinlistoftodos,bordercolor=orange,backgroundcolor=orange!20,linecolor=orange,textsize=scriptsize]{todonotes}
\DeclareMathOperator{\complex}{complexity}
\DeclareMathOperator{\MMD}{MMD}

\DeclareMathOperator{\wass}{W} 
\DeclareMathOperator{\divg}{D} 
\DeclareMathOperator{\gssw}{G_\sigma SW} 
\DeclareMathOperator{\gssww}{G_{\sigma_1} SW}
\DeclareMathOperator{\gsswww}{G_{\sigma_2} SW}

\DeclareMathOperator{\gssd}{G_\sigma SD} 
\DeclareMathOperator{\hatgssd}{\hat{G}_\sigma SD} 

\DeclareMathOperator{\hatgssw}{\hat{G}_\sigma SW} 

\DeclareMathOperator{\gssdd}{G_{\sigma_1} SD}
\DeclareMathOperator{\gssddd}{G_{\sigma_2} SD}

\DeclareMathOperator{\gssmmd}{G_\sigma MMD} 

\usepackage{times}
\usepackage[symbol*]{footmisc}
\DefineFNsymbolsTM{myfnsymbols}{
}%
\setfnsymbol{myfnsymbols}
\allowdisplaybreaks

\usepackage{jmlr}
\begin{document}

\title{Gaussian-Smoothed Sliced Probability Divergences}

\author{
         Mokhtar Z. Alaya 
         \thanks{LMAC EA  2222, Université de Technologie de Compiègne, \emph{e-mail}: \texttt{alayaelm@utc.fr}}
   \and  Alain Rakotomamonjy 
   \thanks{Criteo AI Lab \& LITIS EA4108, Université de Rouen Normandie, \emph{e-mail}: \texttt{alain.rakoto@insa-rouen.fr}}
   \and  Maxime Bérar 
   \thanks{LITIS EA4108, Université de Rouen Normandie, \emph{e-mail}: \texttt{maxime.berar@univ-rouen.fr}}
   \and  Gilles Gasso 
    \thanks{LITIS EA4108, INSA Rouen, \emph{e-mail}: \texttt{gilles.gasso@insa-rouen.fr}}       
}

\maketitle

\begin{abstract}
Gaussian smoothed sliced Wasserstein distance has been recently introduced for comparing probability distributions, while preserving privacy on the data. It has been shown that it provides performances similar to its non-smoothed (non-private) counterpart. However, the computational and statistical properties of such a metric have  not yet been well-established. This work investigates the theoretical properties of this distance as well as those of generalized versions denoted as Gaussian-smoothed sliced divergences $\gssd^p$. We first show that smoothing and slicing preserve the metric property and the weak topology. To study the sample complexity of such divergences, we then introduce $\hat{\hat\mu}_{n}$ the { double empirical distribution} for the smoothed-projected $\mu$. The distribution $\hat{\hat\mu}_{n}$ is a result of a double sampling process: one from sampling according to the origin distribution $\mu$ and the second according to the convolution of the projection of $\mu$ on the unit sphere and the Gaussian smoothing. We particularly focus on the Gaussian smoothed sliced Wasserstein distance $\gssw^p$ and prove that it converges with a rate $O(n^{-1/2})$. We also derive other properties, including continuity, of different divergences with respect to the smoothing parameter. We support our theoretical findings with empirical studies in the context of privacy-preserving domain adaptation.
\end{abstract}


\section{Introduction}
\label{sec:back}

Divergences for comparing two distributions have been shown to be important
for achieving good performance in the contexts  of generative modeling~\citep{arjovsky2017wasserstein,salimans2018improving}, domain adaptation~\citep{long2015learning,courty2016optimal,lee2019sliced}, and in computer vision~\citep{bonnel2011,solomon2015} among many more applications~\citep{klouri17,peyre2019COTnowpublisher,nguyen2023self}.
Examples of divergences that have proved useful for these tasks are the Maximum Mean Discrepancy~\citep{gretton2012kernel, long2015learning, sutherland2017generative}, the Wasserstein
distance~\citep{monge1781,kantorovich1942,villani09optimal} or its variant the sliced Wasserstein distance (SW)~\citep{kolouri2016slicedkernels,bonneel2019spot,kolouri2019gensliced,nguyen2021distributional,nguyen2022hierarchical,nguyen2024markovian}.

The SW distance has the advantage of being computationally
efficient, since it uses a closed-form solution for distributions with support
on $\R$, by computing the expectation of one-dimensional (1D) random projections
of distributions in $\R^d$. Owing to this efficiency and the resulting
scalability, this distance has been successfully
applied in several applications ranging from generative models to domain adaptation~\citep{kolouri2018sliced,max-SW,wu2019sliced,lee2019sliced} 
and its statistical properties have been well-studied in~\cite{nadjahi2020}.

Recently, Gaussian smoothed variants of the Wasserstein distance and 
the sliced Wasserstein distance have been introduced respectively in~\citep{pmlr-v139-nietert21a} and in~\cite{pmlr-v139-rakotomamonjy21a}.
One main motivation behind these variants is to provide a privacy guarantee for the distribution comparison task
as Gaussian smoothing is known to be a mechanism for achieving differential privacy~\citep{dwork2014algorithmic}.
While the properties of the Gaussian smoothed Wasserstein distance have been extensively studied by~\cite{pmlr-v139-nietert21a}, the properties of the Gaussian smoothed sliced Wasserstein distance have not been 
fully investigated yet although they are known to be more computationally efficient.
In this work, we fill this gap by providing a theoretical analysis of the
 Gaussian smoothed sliced Wasserstein distance.

We investigate the theoretical properties of the Gaussian-smoothed sliced Wasserstein distance and the ones of more general Gaussian-smoothed sliced divergences induced by some base distances or divergences for distributions defined in $\R^d$.
Specifically, as for main  contributions, we establish the topological properties of these divergences.  
Then, we focus on the sample complexity of such divergences by introducing the double empirical distribution for the smoothed-projected origin distribution $\mu$. 
The new empirical distribution is a result of double sampling process: one from sampling according to the origin distribution and the second according to the convolution of the projection of $\mu$ on the unit sphere and the Gaussian smoothing. 
We particularly focus on the Gaussian smoothed sliced Wasserstein distance.  Under some mild assumptions we also prove 
 that the Gaussian-smoothed sliced divergences satisfy an order relation with 
 respect to the noise level and are also continuous with respect to this parameter. 
 Given the importance of the noise level in the privacy/utility trade-off achieved by the divergence,
 this latter property is of high impact as it supports
a computationally cheap warm-start/fine-tuning procedure when looking for a 
privacy/utility compromise of the divergence.
Our theoretical study is backed by some numerical experiments on 
toy problems and on 
domain adaptation illustrating how owing to the
topology induced by our metric and its continuity,  differential privacy comes almost 
for free (without loss of performance) and
multiple models with different level of privacy can be cheaply computed.

\paragraph{Comparison with previous works.}

Here we highlight the position of this work compared to the most linked previous ones, in particular 
\citet{nadjahi2020} and \citet{pmlr-v139-rakotomamonjy21a}. The work of~\citet{nadjahi2020} is focused on sliced Wasserstein distance and its statistical properties, however our work is based on the properties of the Gaussian smoothed with general divergences (e.g. Wasserstein, MMD, Sinkhorn divergence).
We argue that the properties cannot be directly derived from~\citep{nadjahi2020}, especially the sample complexity result.
In~\citet{pmlr-v139-rakotomamonjy21a}, the authors investigated the smoothed Wasserstein distance and their theoretical finding was principally on proving the metric property, whereas we further investigate sample and projection complexities and the  continuity properties w.r.t. the smoothing noise level.  

We emphasize that the novelty of the present paper consists in 
the theoretical properties derived from the definition of the empirical measure $\hat{\hat\mu}_{n}$. This latter is derived from a double process sampling, which is inspired from the implementation part. Namely, we sample $X_1, \ldots, X_n$ from the raw distribution $\mu$ to define $\hat\mu_n$ then project it on the unit sphere and smooth this projection with a Gaussian distribution. Nevertheless, this smoothing, from a theoretical point view, is a continuous measure (see Lemma~\ref{lem:law_of_sum_of_smooth_projections}) that needs to be sampled. This entails our introducing to the second sampling step and construct $\hat{\hat\mu}_{n}$, an empirical version for the smoothing projection of $\mu$. To the best of our knowledge, this work is the first introducing the double randomness in the case of smoothing optimal transport discrepancies. Recent works~\citep{Goldfeld-IEEE20,pmlr-v139-nietert21a} addressed the smoothing Wasserstein an their theoretical results relied only on $\hat \mu_n$.

\paragraph{Layout of the paper.}
The paper is organized as follows: after introducing the 
notation and some background in Section \ref{sec:back}, we detail
the topological properties of Gaussian-smoothed sliced divergence in 
Section \ref{subsec:topo} while the {double sampling process} and its statistical properties are established in Section
\ref{subsec:stats}. The noise analyses are provided in Section \ref{subsec:noise}. 
 Experimental analyses for supporting the theory and 
showcasing the relevance of our divergences in domain adaptation
are depicted in Section \ref{sec:expe}. Discussions on the perspectives
and limitations  are in Section \ref{sec:conclu}. All the proofs of the theoretical results and some additional experiments are postponed to the appendices in the supplementary.


\section{Preliminaries}
\label{sec:back}

For the reader's convenience, we provide a brief summary of standard notations and definitions used throughout the paper.

\paragraph{Notation.} For $d \in \mathbb{N}^*$,   let $\cP(\R^d)$ be the set of Borel probability measures on  $\R^d$ and $\cP_p(\R^d) \subset \cP(\R^d)$, those with finite moment of
order $p$, {i.e.,} $\cP_p(\R^d) \triangleq \{\mu \in \cP : \int \|x\|^p d\mu(x) < \infty\}$, where $\|\cdot\|$ is the Euclidean norm. 
We denote $M_p(\mu) = \int_{x} \norm{x}^p \diff \mu(x).$
For two probability distributions $\mu$ and $\nu$, we denote their convolution as
$\mu * \nu \in \cP(\R^d)$, namely
$(\mu * \nu)(A) = \int_x\int_y \mathbf{1}_A(x+y) \diff\mu(x) \diff\nu(y)$, where 
$\mathbf{1}_A(\cdot)$ is the indicator function over $A$. Given two independent random
variables $X \sim \mu$ and $Y \sim \nu$, we remind that $X + Y \sim \mu * \nu$.
The $d$-dimensional unit-sphere is noted as $\mathbb{S}^{d-1} \triangleq \{\theta \in \R^d : \|\theta\| = 1 \} $. We denote by $u_d$ the uniform distribution on $\mathbb{S}^{d-1}$ and we use $\delta(\cdot)$ to denote the Kronecker delta function. 
We note as $\E_\mu f$ the expectation of the function $f$ with respect to $\mu$. 

Let $\Gamma:\R \rightarrow \R$ be the Gamma function expressed as $\Gamma(v) = \int_0^\infty t^{v-1}e^{-t}dt$ for $v>0$. For $k \in \mathbb N$, 
 $(\cdot)_k$ denoted the Pochhammer symbol, also known in the literature
as a rising factorial, namely  $(\alpha)_k = \frac{\Gamma(\alpha +k)}{\Gamma(k)} = \alpha(\alpha+1)\cdots (\alpha +k -1)$.
We denote by ${}_1F_1(\alpha, \gamma; z)$ the Kummer’s confluent hypergeometric function~\citep{olver2010nist} and defined by ${}_1F_1(\alpha, \gamma; z) = \sum_{k=0}^\infty \frac{(\alpha)_k}{(\gamma)_k} \frac{z^k}{k!}.$

\paragraph{Sliced Wasserstein distance.} We remind in this paragraph several measures of similarity between two distributions.
The Wasserstein distance of order $p \in [1, \infty)$ between two measures in $\mathcal{P}_p(\R^d)$ is given by the
relaxation of the optimal transport problem,
and it is defined as
\begin{equation*}\label{eq:wd}
\wass_p^p(\mu,\nu) = \inf_{\gamma \in \Pi(\mu,\nu)} \int_{\R^d \times \R^d} 
\|x - x^\prime\|^p \gamma(x,x^\prime) \diff x\diff x^\prime
\end{equation*} 
where $\Pi(\mu,\nu)\triangleq \{ \gamma \in \mathcal{P}(\R^d \times \R^d) |  \pi_{1\#} \gamma=\mu,\pi_{2\#} \gamma=\nu\}$ and $\pi_1, \pi_2$ are
the marginal projectors of $\gamma$ on each of its coordinates. 
When $d=1$, the Wasserstein distance can be calculated in closed-form  owing to the
cumulative distributions of $\mu$ and $\nu$~\citep{rachev1998mass}.
 In practice for empirical distributions, the closed-form solution requires only the sorting of the samples, which makes it very efficient. Because of this efficiency, efforts have been devoted to derive a metric for high-dimensional distributions based on 1D Wasserstein distance.
The main idea is to project high-dimensional probability distributions onto a random one-dimensional space and
then to compute the Wasserstein distance. This  operation
can be theoretically formalized through the use of the Radon transform, leading to the so-called sliced Wasserstein distance~\citep{kolouri2016slicedkernels,bonneel2019spot,kolouri2019gensliced,nguyen2021distributional}.

\begin{definition}  For any $p \in [1, \infty)$ and two measures
	$\mu$, $\nu\in \mathcal{P}_p(\R^d)$, the sliced Wasserstein distance (SW) reads as 
\begin{equation*}
\text{SW}^p(\mu,\nu) \triangleq \int_{\mathbb{S}^{d-1}}  \wass_p^p(\mathcal{R}_\u \mu,\mathcal{R}_\u \nu) u_d(\u)\diff\u. \end{equation*}
where $\mathcal{R}_\u$ is the Radon transform of a probability distribution, namely
$\mathcal{R}_\u \mu(\cdot) = \int_{\R^d} \mu(\s) \delta(\cdot - \s^\top \u)d\s\label{eq:radon}$.
In practice, the integral is approximated through a Monte-Carlo simulation leading to a sum of 1D Wasserstein distances over a fixed number of random directions $\u$.  
\end{definition}

\paragraph{Gaussian-smoothed sliced Wasserstein distance.} 

Based on this definition of SW, replacing the Radon projected measures with their Gaussian-smoothed counterpart leads to the following definition:
\begin{definition}
	The $\sigma$-Gaussian-smoothed $p$-Sliced Wasserstein distance between probability distributions $\mu$ and $\nu$ in  $\mathcal{P}_p(\R^d)$ writes as 
	\begin{equation*}
		\label{eq:gsmoothedwass}
	\gssw^p(\mu,\nu) \triangleq\int_{\mathbb{S}^{d-1}} \!\!\! \wass_p^p(\mathcal{R}_\u \mu * \mathcal{N}_\sigma,\mathcal{R}_\u \nu* \mathcal{N}_\sigma) u_d(\u)\diff\u, 	\end{equation*}
\end{definition}
where $\mathcal{N}_\sigma = \mathcal{N} (0, \sigma^2)$ is the zero-mean $\sigma^2$-variance Gaussian measure.
It is important to note here that the smoothing (convolution) operation occurs after projection onto the one-dimensional space. Hence, assuming $X \sim \mu$, $Y \sim \nu$, for a given direction $\u$, we compute in the integral the one-dimensional Wasserstein distance  between the probability laws of $\u^\top X + Z$ and $\u^\top Y + Z^\prime$ where $Z,Z^\prime \sim \mathcal{N}_{\sigma}$ are independent random variables. 
{The metric properties of $\gssw_p$ for $p \geq 1$ have been discussed in a recent work~\citep{pmlr-v139-rakotomamonjy21a}. 
This latter work has also shown, in the context of differential privacy, 
the importance of convolving the Radon projected distribution with a
Gaussian instead of computing the SW distance of the original distribution smoothed with a $d$-dimensional Gaussian {$\mu * \mathcal{N}_{\sigma \mathbf{I}_d}$, } where $\mathbf{I}_d$ denotes the $d\times d$ identity matrix.}

\paragraph{Gaussian-smoothed sliced divergence.}  
The idea of slicing high-dimensional distributions before feeding them to a divergence between probability distributions can be extended
to distances other than the Wasserstein distance. These sliced
divergences have been studied by~\citet{nadjahi2020}. Similarly, 
we can define a Gaussian-smoothed  sliced divergence, given a 
divergence $\divg_{\R^d} : \cP_p(\R^d) \times \cP_p(\R^d) \rightarrow \R^+$ for $d \geq 1$ as:
\begin{definition} \label{def:Gaussian_smoothed_sliced_Div}
	The $\sigma$-Gaussian-smoothed $p$-Sliced Divergence between probability distributions $\mu$ and $\nu$ in $\cP_p(\R^d)$
	associated to the {\it base divergence}  $\divg \triangleq \divg_\R$, $p\geq 1$  is 
		\begin{equation*}
		\label{eq:gsmootheddiv}
\gssd^p(\mu,\nu) \triangleq \int_{\mathbb{S}^{d-1}}  \divg^p(\mathcal{R}_\u \mu * \mathcal{N}_\sigma,\mathcal{R}_\u \nu* \mathcal{N}_\sigma) u_d(\u)\diff\u. \end{equation*}    
where the superscript $p$ refers to a power.
\end{definition}

Typical relevant divergences are the maximum mean discrepancy (MMD) \citep{gretton2012kernel} or the Sinkhorn divergence \citep{genevay2018learning,peyre2019COTnowpublisher}. In Section \ref{sec:expe}, we report empirical findings based on these divergences as well as on the Wasserstein distance.


\section{Theoretical properties}
\label{sec:theory}
In this section, we analyze the properties of 
the Gaussian-smoothed sliced divergence, in terms of
topological and statistical properties and the influence
of the Gaussian smoothing parameter $\sigma$ on the distance.

\subsection{Topology} \label{subsec:topo}

It has already been shown in~\cite{pmlr-v139-rakotomamonjy21a} that the Gaussian-smoothed sliced 
Wasserstein is a metric on $\cP(\R^d)$. In the next, we extend these results to any divergence $\divg(\cdot,\cdot)$  under certain assumptions.

\begin{theorem} 
\label{theorem:proof_topology} For any $\sigma>0, p\geq 1$,  the following properties hold:
	\begin{enumerate}			\item if $\divg(\cdot,\cdot)$ is non-negative (or symmetric), then $\gssd(\cdot,\cdot)$ is non-negative (or symmetric);
\item if $\divg(\cdot,\cdot)$ satisfies
the identity of indiscernibles, i.e. for $\mu^\prime, \nu^\prime \in \cP(\R) $, $\divg(\mu^\prime, \nu^\prime) = 0$ if and only if 
$\mu^\prime = \nu^\prime$, then this identity also holds for $\gssd(\cdot,\cdot)$ for any $\mu,\nu \in \mathcal{P}_p(\R^d)$;
\item if $\divg(\cdot,\cdot)$ satisfies the triangle inequality then 
$\gssd(\cdot,\cdot)$ satisfies the triangle inequality.
	\end{enumerate}

\end{theorem}

The above theorem shows that under mild hypotheses over the base divergence $\divg$, as being a metric for instance, the metric property of its Gaussian-smoothed sliced version naturally derives. As exposed in the appendix, the more involved property to prove is the identity of indiscernibles.

We further postponed to the appendix the proofs of the two other topological properties: (i) $\gssd$ metrizes the weak topology on $\mathcal{P}_p(\R^d)$ and (ii) $\gssd$ is  lower semi-continuous with respect to  the weak topology in $\mathcal{P}_p(\R^d)$.

Now, we establish under which  conditions on the divergence $\divg$, the convergence of a sequence in
$\gssd$ implies weak convergence in $\cP_p(\R^d)$. We say that $\{\mu_k\}_{k \in \mathbb N}$ {\it converges weakly} to $\mu$ and write, $\mu_k \Rightarrow\mu$, if $\int f(x) d \mu_k(x) \rightarrow
\int f(x) d \mu(x)$, as $k\rightarrow \infty,$
for every $f$ in the space of all bounded continuous real functions.

\begin{theorem} 
\label{theorem-weaktopo}
Let $\sigma > 0, p \geq 1$, $\mu
\in \cP_p(\R^d)$, and $\{\mu_k\in \cP_p(\R^d)\}_{k\in \mathbb N}$  a sequence of distributions. Assume that the divergence $\divg$ is bounded and metrizes the weak topology on $\cP(\R)$. 
Then, $ \lim_{k \rightarrow \infty }\gssd(\mu_k,\mu) = 0 $ if and only if $\mu_k \Rightarrow\mu.$
	
\end{theorem}

Note that Theorem~\ref{theorem-weaktopo} extends the results of~\cite{nadjahi2020} 
to Gaussian-smoothed distributions, as we retrieve them as a special case for $\sigma = 0$. In addition, based on Theorem
3.2 by \cite{lin2021projection} and the above, we can also claim  that
the Gaussian-smoothed SWD metrizes the weak convergence.

\begin{proposition}
\label{prop_lower_semicont}
Let $\sigma>0, p\geq 1$ and assume that the base divergence $\divg$ is lower semi-continuous w.r.t.  the weak topology in $\mathcal{P}({\R})$.
Then, $\gssd$ is  lower semi-continuous with respect to  the weak topology in $\mathcal{P}_p(\R^d)$. 
\end{proposition}
{When the base divergence {$\divg$ }	is equal to the Wasserstein distance $\wass_p$, 
that is lower semi-continuous~\citep{villani09optimal}, then Proposition~\ref{prop_lower_semicont} 
shows that the smoothed sliced Wasserstein distance is semi-lower continuous too.


\subsection{Statistical properties} \label{subsec:stats}

The next theoretical question we are interested in is about the incurred error
 when the true distribution $\mu$ is approximated by its empirical distribution $\hat \mu_n$. Such a case is common in practical applications where only (high-dimensional) empirical samples are at disposal. 
 Specifically, we are interested in quantifying two key properties of empirical Gaussian-smoothed
 divergence: {\it (i)} 
 the convergence of the double empirical 
 $\hatgssd( {\hat\mu}_n,{\hat\nu}_n)$ (see Definition~\ref{def:double_randomness})  to $\gssd(\mu,\nu)$
{\it (ii)} 
the convergence of 
$ \widehat{\gssd} (\mu,\nu)$ (see \eqref{eq:MC_empirical}) to $\gssd(\mu,\nu)$,  when
approximating the expectation over the random projection with sample mean.

Let $\hat \mu_n = \frac1n \sum_{i=1}^n \delta_{X_i}$ and $\hat \nu_n = \frac1n \sum_{i=1}^n \delta_{Y_i}$ be the empirical probability measures of independent observations. The smoothed Gaussian sliced divergence between $\hat \mu_n$ and $\hat \nu_n$ is given by
\begin{align*}
{\gssd^p} &(\hat \mu_n,\hat \nu_n) =\int_{\mathbb{S}^{d-1}} \divg^p\big(\mathcal{R}_\u \hat\mu_n * \mathcal{N}_\sigma, \mathcal{R}_\u \hat\nu_n * \mathcal{N}_\sigma\big)u_d(\u)\diff\u.
\end{align*}
\begin{remark}
Remark that for a fixed $\u \in {\mathbb{S}^{d-1}}$, the distributions $\mathcal{R}_\u \hat\mu_n * \mathcal{N}_\sigma$ and $\mathcal{R}_\u \hat\nu_n * \mathcal{N}_\sigma$ are {\it continuous}, in particular they are a mixture of Gaussian distributions centered on the projected samples with variance $\sigma^2.$
\end{remark}
\begin{lemma}
\label{lem:law_of_sum_of_smooth_projections}
Conditionally on the samples $\{X_i\}_{i=1, \ldots, n}$ and $\{Y_i\}_{i=1, \ldots, n}$, one has:
$\mathcal{R}_\u \hat\mu_n * \mathcal{N}_\sigma = \frac 1n \sum_{i=1}^n \mathcal{N}(\u^\top X_i, \sigma^2)$
and 
$\mathcal{R}_\u \hat\nu_n * \mathcal{N}_\sigma = \frac 1n \sum_{i=1}^n \mathcal{N}(\u^\top Y_i, \sigma^2).$
\end{lemma}

Note that 
we further need to sample with respect to the continuous mixture Gaussian measures in Lemma~\ref{lem:law_of_sum_of_smooth_projections} 
in order to get a {\it fully } empirical measure version of $\gssd(\mu,\nu)$. To this end, we next define the {\it double} empirical divergence of $\gssd.$

\subsubsection{Double empirical divergence of $\gssd$}

Let $T^x_1, \ldots, T^x_n$ and $T^y_1, \ldots, T^y_n$  be i.i.d. observations of $\mathcal{R}_\u \hat\mu_n * \mathcal{N}_\sigma  
 $ 
 and $\mathcal{R}_\u \hat\nu_n * \mathcal{N}_\sigma ,$
respectively. Sampling i.i.d. $\{T^x_i\}_{i=1, \ldots, n}$  is given by the following scheme:
for $ i = 1, \ldots, n$, we first choose the component $\mathcal{N}(\u^\top X_i, \sigma^2)$ from the mixture $\frac 1n \sum_{i=1}^n \mathcal{N}(\u^\top X_i, \sigma^2)$ then we generate $T^x_{i} = \u^\top X_i + Z^x_i$, where $Z^x_i \sim \mathcal{N}_\sigma.$
Hence, we set, for a given $\u$ 
\begin{align*}
\hat{\hat\mu}_{n} 
= \frac 1n \sum_{i=1}^n \delta_{T_{i}^x} = \frac 1n \sum_{i=1}^n \delta_{\u^\top X_i + Z^x_i} \text{ and } \hat{\hat\nu}_{n} 
= \frac 1n \sum_{i=1}^n\delta_{T_{i}^y} = \frac 1n \sum_{i=1}^n \delta_{\u^\top Y_i + Z^y_i}. 
\end{align*}
The measure $\hat{\hat\mu}_{n} \in \mathcal{P}(\R)$ defines an empirical version of the continuous $\mathcal{R}_\u \hat\mu_n * \mathcal{N}_\sigma$ denoted as  
$\widehat{\mathcal{R}_\u \hat\mu_n * {\mathcal{N}_{\sigma}}}$ (similarly $\hat{\hat\nu}_{n} = \widehat{\mathcal{R}_\u \hat\nu_n * {\mathcal{N}_{\sigma}}}$).
Using the aforementioned notation, we define. 
\begin{definition}
\label{def:double_randomness}
The double empirical smoothed Gaussian sliced divergence reads as 
\begin{equation*}
{\hatgssd^p} (\hat \mu_n,\hat \nu_n) \triangleq \int_{\mathbb{S}^{d-1}} \divg^p(\hat{\hat\mu}_{n}, \hat{\hat\nu}_{n} )u_d(\u)\diff\u.\end{equation*}
\end{definition}

\begin{remark}
{\it (i)} It is worth to comment the double randomnesses showing in the  definition of $\hatgssd^p (\hat \mu_n,\hat \nu_n)$: the first comes from sampling according to the original probability measure ($\mu$ or $\nu$) whereas the second takes place from sampling according to the mixture $\frac 1n \sum_{i=1}^n \mathcal{N}(\u^\top X_i, \sigma^2)$. \\
{\it (ii)} The empirical measure of the convolution $\widehat{\mathcal{R}_\u \mu * {\mathcal{N}_{\sigma}}}$ could be written as $\frac 1n \sum_{i=1}^n \delta_{U^x_i + Q^x_i}$ allowing to sample {\it in a one shot} $n$ i.i.d. samples $U^x_i + Q^x_i$ such that $U^x_i \sim \mathcal{R}_\u \mu$ and $Q^x_i \sim \mathcal{N}_\sigma$. 
From an empirical view, sampling according to ${\mathcal{R}_\u \mu * {\mathcal{N}_{\sigma}}}$ is intractable. For that reason, our theoretical results and numerical experiments are based on 
$\hat{\hat\mu}_{n}, \hat{\hat\nu}_{n}$, 
and hence with respect to ${\hatgssd} (\hat \mu_n,\hat \nu_n).$ 


\end{remark}

\subsubsection{Sample complexity of $\gssw^p$}

Herein, our goal is to quantify the error made when approximating $\gssw(\mu,\nu)$ with ${\hatgssw} (\hat{\mu}_n,\hat{\nu}_n)$. 
More precisely, we are interested in establishing an order of the convergence rate of  ${\hatgssd} (\hat{\mu}_n,\hat{\nu}_n)$ towards $\gssd(\mu,\nu)$, according to the sample size $n.$ This rate stands for the so-called {\it sample complexity.} 

{
The convergence results in the sequel are given in expectation. Recall that the empirical distributions are derived from a double sampling process, which leads to consider a double expectations, wrt the origin distribution $\E_{\mu^{\otimes_n}}$ and wrt the sampling from the Gaussian smoothing $\E_{\mathcal{N}^{\otimes_n}_\sigma}$ where  $\mu^{\otimes_n}$ and  $\mathcal{N}^{\otimes_n}_\sigma$ are the $n$-fold product extensions of $\mu$ and $\mathcal{N}_\sigma$, respectively. We first consider the conditional expectation given the samples $X_1, \ldots, X_n$, i.e. $\E_{\mathcal{N}^{\otimes_n}_\sigma}[\cdot | X_1, \ldots, X_n]$, and then apply   $\E_{\mu^{\otimes_n}}$. We denote by 
\begin{align*}
\E_{\mu^{\otimes_n}| \mathcal{N}^{\otimes_n}_\sigma} [\cdot] = \E_{\mu^{\otimes_n}}\big[\E_{\mathcal{N}^{\otimes_n}_\sigma}[\cdot | X_1, \ldots, X_n]\big].
\end{align*}
Next, we focus on the sample complexity for the special case of Gaussian-smoothed sliced Wasserstein distance. 
}


\begin{proposition}  
\label{theorem:sample_double_empirical_complexity}
Fix $\sigma > 0, p\geq 1$ and $\vartheta > \sqrt{2}$. For $X\sim \mu$, assume that $\int_{0}^\infty e^{\frac{2\xi^2}{\sigma^2\vartheta^2}}\P\big[\norm{X} > \xi\big]\diff \xi < \infty.$
Then, 
\begin{align*}
\E_{\mu^{\otimes_n}| \mathcal{N}_\sigma^{\otimes_n}}[{\hatgssw^p}({\hat\mu}_n,\mu)]
\leq \Xi_{p,\sigma, \vartheta} \frac{1}{\sqrt{n}}+ \Upsilon_{p,\sigma,\vartheta, \mu} \frac{\log n}{n},
\end{align*}
where $\Xi_{p,\sigma, \vartheta} = \frac{2 ^{\frac{5p}{2} - \frac 54}}{\sqrt{\pi}} 
\sigma^{p- \frac 14} \vartheta^{p+1}\sqrt{\Gamma\big(p+ \frac 12\big)}\Big(\sqrt{\frac{4 \pi \sigma^2\vartheta^2}{\vartheta^2 - 2}} + 4 \int_{0}^\infty e^{\frac{2\xi^2}{\sigma^2\vartheta^2}}\P\big[\norm{X} > \xi\big]\diff \xi\Big)^{\frac 12}$ and 
$\Upsilon_{p,\sigma,\vartheta, \mu} = \frac{2^{2p-1}C_{p}}{\sqrt{\pi}}\sigma^{2p} \Gamma(p + \frac 12) {}_1F_1(-p, \frac 12; M_{2k}(\mu))$
with $C_p$ is a positive constant depending only on $p$.
 \end{proposition}
It is worth to note that for  $p \in \mathbb N^*$, e.g. $p=2$  (standard choice for numerical experiments), the confluent hypergeometric function ${}_1F_1(-p, \frac 12; M_{2k}(\mu))$ becomes a polynomial since $(-p)_{(k)}=0$ for $k\geq p+1.$  Now, let us sketch the proof of Proposition~\ref{theorem:sample_double_empirical_complexity}: we first insert the proxy term of mixture Gaussian distribution $\frac 1n \sum_{i=1}^n\mathcal{N}(\u\top X_i, \sigma^2)$, then by an application of the triangle inequality on the Wasserstein distance we are faced to control two terms (i) $\wass_p^p(\hat{\hat \mu}_n, \frac 1n \sum_{i=1}^n\mathcal{N}(\u\top X_i, \sigma^2))$ and (ii) $\wass_p^p(\frac 1n \sum_{i=1}^n\mathcal{N}(\u\top X_i, \sigma^2),\mu)$. For (i)  we get a standard order of $O(\frac{\log n}{n})$, which  comes from a by-product of~\cite{fournier2015}.
For (ii), through a coupling via the maximal coupling using the total variation distance (Theorem 6.15 in \cite{villani09optimal}), we obtain the order $O(n^{-1/2})$. The control technique for (ii) was inspired from~\cite{Goldfeld-IEEE20} and~\cite{pmlr-v139-nietert21a}.

\begin{remark}
The condition $\int_{0}^\infty e^{\frac{2\xi^2}{\sigma^2\vartheta^2}}\P\big[\norm{X} > \xi\big]\diff \xi < \infty$ needs $\P\big[\norm{X} > \xi\big]$ goes to $0$ faster than $e^{-\kappa\xi^2}$ for $\kappa < 2/\sigma^2\vartheta^2$. This can be satisfied when $\norm{X}$ is a $\omega$-sub-gausssian ($\omega\geq 0)$. Namely,  $\E[e^{\eta^\top(X - \E[X])}] \leq e^{\frac{\omega \norm{\eta}^2}{2}}$ for all $\eta \in \R^d.$ If the parameter $\omega$ verifies $\omega < \sigma \vartheta /2$, then the latter condition holds.
\end{remark}

\begin{remark}
Note that the sample complexity depends
on the amount of smoothing through the moment of the Gaussian
noise : the larger the amount of smoothing (and thus the privacy), the worse is the constant of the complexity. Hence, a 
trade-off on privacy and statistical estimation appears
here as a reasonable guarantee on  the differential privacy usually  requires a large Gaussian variance. 
\end{remark}

\begin{proposition} 
\label{theorem:first_control_sample_complexity_for_mu_and_nu}
Under the same conditions of Proposition~\ref{theorem:sample_double_empirical_complexity}, we have 
\begin{align*}
\E_{\mu^{\otimes_n}| \mathcal{N}_\sigma^{\otimes_n}}
\E_{\nu^{\otimes_n}| \mathcal{N}_\sigma^{\otimes_n}}[{\hatgssw^p}({\hat\mu}_n,\hat\nu_n)]\leq 3^{p-1} {\gssw^p}({\mu},\nu) +  3^{p}\Xi_{p,\sigma, \vartheta} \frac{1}{\sqrt{n}} + 3^{p-1}(\Upsilon_{p,\sigma,\vartheta, \mu} + \Upsilon_{p,\sigma,\vartheta, \nu}) \frac{\log n}{n}
\end{align*}
and 
\begin{align*}
{\gssw^p}({\mu},\nu)
\leq 3^{p-1} \E_{\mu^{\otimes_n}| \mathcal{N}_\sigma^{\otimes_n}}
\E_{\nu^{\otimes_n}| \mathcal{N}_\sigma^{\otimes_n}}[{\hatgssw^p}({\hat\mu}_n,\hat\nu_n)] +  3^{p}\Xi_{p,\sigma, \vartheta} \frac{1}{\sqrt{n}} + 3^{p-1}(\Upsilon_{p,\sigma,\vartheta, \mu} + \Upsilon_{p,\sigma,\vartheta, \nu}) \frac{\log n}{n}
\end{align*}

\end{proposition}

{Proof of Proposition~\ref{theorem:first_control_sample_complexity_for_mu_and_nu} relies on a double application of triangle inequality satisfied by Wasserstein distance as follows: ${\wass_p}(\hat{\hat\mu}_{n},\hat{\hat\nu}_{n}) \leq {\wass_p}(\hat{\hat\mu}_{n}, \mathcal{R}_\u \mu* \mathcal{N}_\sigma) + {\wass_p}(\mathcal{R}_\u \mu* \mathcal{N}_\sigma, \mathcal{R}_\u \nu* \mathcal{N}_\sigma) + {\wass_p}(\mathcal{R}_\u \nu* \mathcal{N}_\sigma, \hat{\hat\nu}_{n})$,
combined with Proposition~\ref{theorem:sample_double_empirical_complexity}. This gives a {non sharp} convergence result since we get the constant $3^{p-1}$ in front of $\E_{\mu^{\otimes_n}| \mathcal{N}_\sigma^{\otimes_n}}
\E_{\nu^{\otimes_n}| \mathcal{N}_\sigma^{\otimes_n}}[{\hatgssw^p}({\hat\mu}_n,\hat\nu_n)]$ or ${\gssw^p}({\mu},\nu)$. However, when the power $p=1$ we obtain a sharp convergence result with $O(n^{-1/2})$, namely
\begin{align*}
|\E_{\mu^{\otimes_n}| \mathcal{N}_\sigma^{\otimes_n}}
\E_{\nu^{\otimes_n}| \mathcal{N}_\sigma^{\otimes_n}}[{\hatgssw}({\hat\mu}_n,\hat\nu_n)]-  {\gssw}({\mu},\nu)| \leq 3^{}\Xi_{1,\sigma, \vartheta} \frac{1}{\sqrt{n}}.
\end{align*}
}
Despite that our theoretical results hold only for Gaussian-smoothed sliced Wasserstein distance, our empirical results show that given other base divergences $\divg$, shows that the sample complexity of $\gssd^p$ is proportional to the one dimensional sample complexity of $\divg^p$ ($p=2)$. {Figure \ref{fig:divsamples} provides an empirical illustration of this statement.}

\begin{figure}[t]
	\centering
	\includegraphics[width=5cm]{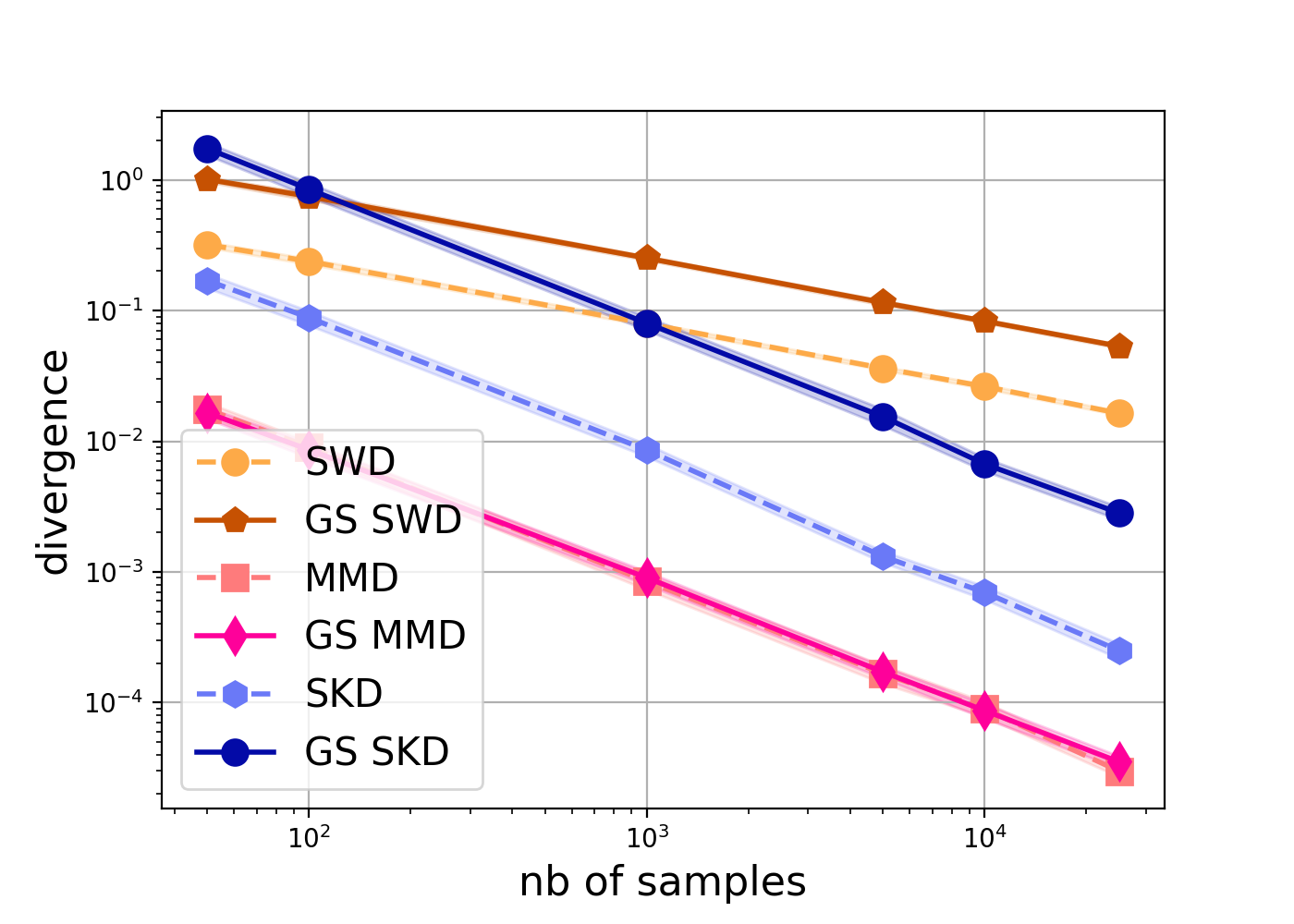}
	\includegraphics[width=5cm]{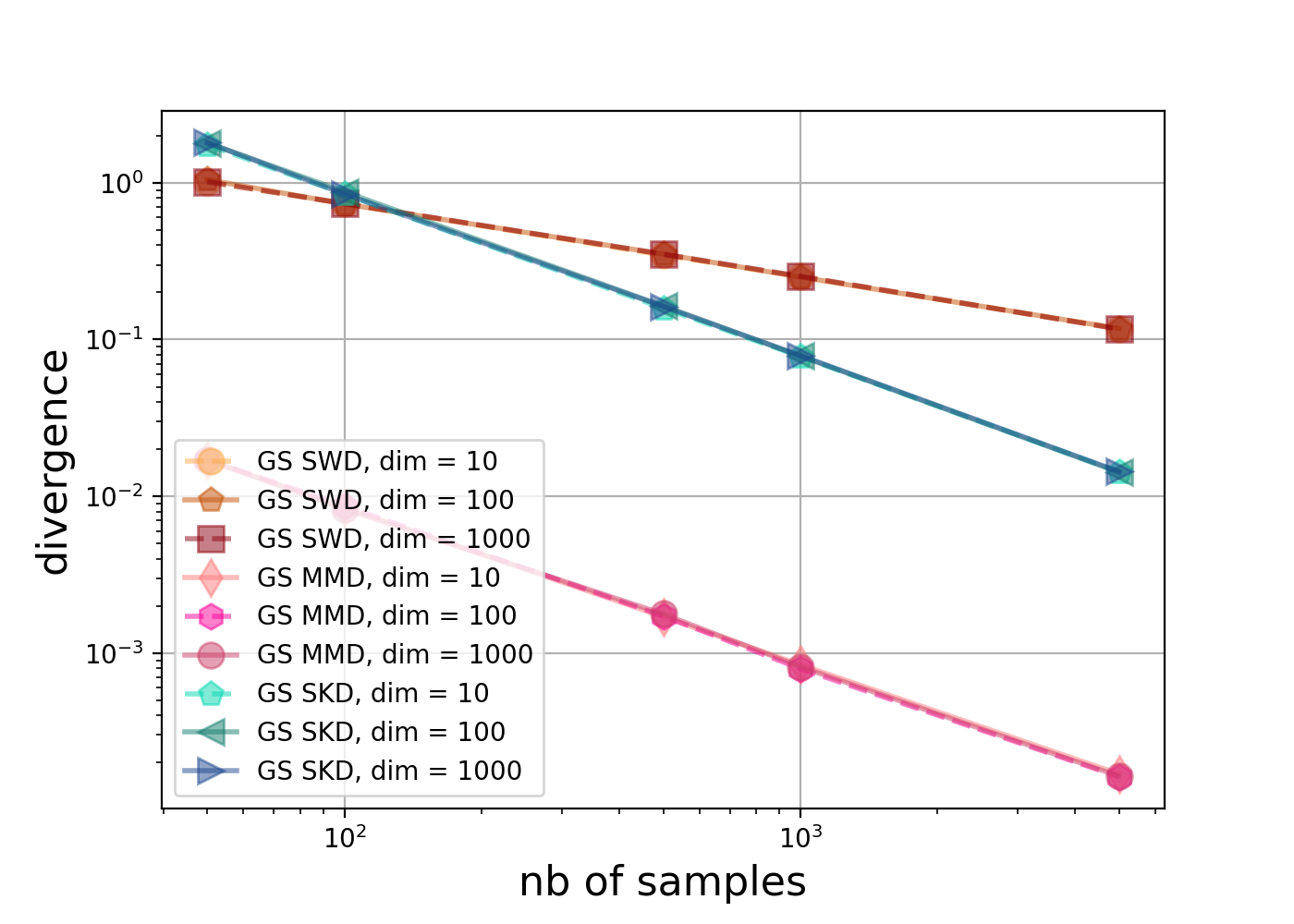}

	\caption{
	Measuring the divergence between two sets of samples in $\R^{50}$, of increasing size,
		randomly drawn from $\mathcal{N}(0,\mathbf{I})$.
		We compare three sliced divergences and their Gaussian-smoothed sliced versions
		with a $\sigma=3$: (top) dimension has been set to $d=50$; (bottom) sample complexity       with different dimensions. This plot confirms that the complexity is dimension-independent.
		\label{fig:divsamples}}  
\end{figure}



\subsubsection{Projection complexity}

To compute the Gaussian-smoothed sliced divergence, one may resort to  a Monte Carlo scheme to numerically approximate the integral in $\gssd^p(\mu,\nu)$. Towards this, let define the following sum:
\begin{align}
\label{eq:MC_empirical}
\widehat{\gssd^p}(\mu,\nu) = 
\frac{1}{L}
\sum_{l=1}^L 
\divg^p(\mathcal{R}_{\u_l}\mu * \mathcal{N}_\sigma,\mathcal{R}_{\u_l} \nu * \mathcal{N}_\sigma),
\end{align}
where $\u_l$ is a random vector uniformly drawn  from $\mathbb{S}^{d-1}$, for $l=1, \ldots, L.$ Theorem~\ref{theorem:error_MC} shows that for a fixed dimension $d$, the root mean square error of Monte Carlo (MC) approximation is of order $O\big(\frac{1}{\sqrt{L}}\big)$, which corresponds to the projection complexity.
We denote by $u_d^{\otimes_L}$ and the $L$-fold product extensions of the uniform measure $u_d$ on the unit sphere.
\begin{proposition}
\label{theorem:error_MC}
Let $\sigma > 0, p\geq 1$. Then
the error related to the MC-estimation of $\gssd^p$ is bounded as follows
\begin{align*}
\E_{u_d^{\otimes_L}}[|\widehat{\gssd^p}( \mu,\nu) - \gssd^p(\mu,\nu)|] \leq \frac{A(p,\sigma)}{\sqrt{L}},
\end{align*}
where $A^2(p,\sigma) = \int_{\mathbb{S}^{d-1}}\big(\divg^p(\mathcal{R}_{\u}\mu * \mathcal{N}_\sigma,\mathcal{R}_{\u} \nu* \mathcal{N}_\sigma) - \bar\vartheta_p\big)^2u_d(\u)\diff\u,$
with $\bar\vartheta_p = \int_{\mathbb{S}^{d-1}}\divg^p(\mathcal{R}_{\u}\mu * \mathcal{N}_\sigma,\mathcal{R}_{\u} \nu* \mathcal{N}_\sigma)u_d(\u)\diff\u$.
\end{proposition}
The term $A^2(p,\sigma)$ corresponds to the variance of $\divg^p(\mathcal{R}_{\u}\mu * \mathcal{N}_\sigma,\mathcal{R}_{\u} \nu* \mathcal{N}_\sigma)$ with respect to $\u \sim u_d$. 
It is worth to note that the precision of the Monte Carlo scheme approximation depends on the number of projections $L$ and the variance of the evaluations of the divergence $\divg^p.$ The estimation error decreases at the rate $L^{-1/2}$ according to  the number of projections used to compute the smoothed sliced divergence. 
Given the above results, we provide a finer analysis of $\gssw^p(\mu,\nu)$'s sample complexity. %
{
\begin{corollary}
The sample and projection complexities of $\gssw^p(\mu,\nu)$ reads as $\complex(\gssw^p) = O(n^{-1/2} + L^{-1/2}).$ If we consider the number of projections as $L = \lfloor n^{\beta}\rfloor $ for some $\beta \in (0, 1)$ then the overall complexity $\complex(\gssw^p(\mu,\nu)) = O(n^{-\beta/2})$. 
\end{corollary}
}



\subsection{Noise-level dependencies} \label{subsec:noise}

The parameter $\sigma$ of the Gaussian smoothing function $\mathcal{N}_{\sigma}$ may 
significantly influence the attained privacy level. 
Hence, we provide theoretical results analyzing the effect of 
the noise level $\sigma$ on the induced Gaussian-smoothed sliced divergence.

\subsection{Order relation.}

We first show that the noise level tends to reduce 
the difference between two distributions as measured using 
$\gssd^p(\mu,\nu)$ provided the base divergence $D$
satisfies some mild assumptions.

\begin{proposition}
\label{proposition:2-level-noise}
 Let $\mu, \nu \in \cP_p(\R^d)$  and 
 consider the noise levels
    $\sigma_1, \sigma_2$ such that 
    $0 \leq \sigma_1 \leq \sigma_2 < \infty$. Assume that the base divergence $\divg$ satisfies $\divg(\mu'* \mathcal{N}_{\sigma_2}, \nu'* \mathcal{N}_{\sigma_2}) \leq 
   \divg(\mu' * \mathcal{N}_{\sigma_1}, \nu' * \mathcal{N}_{\sigma_1}),$
            for any $\mu', \nu' \in \mathcal{P}(\R).$ Then,  $\gssddd^p(\mu,\nu) \leq \gssdd^p(\mu,\nu).$
\end{proposition}
Note that the assumption for the base divergence inequality holds for the Gaussian-smoothed Wasserstein distance~\cite{pmlr-v139-nietert21a}. While we conjecture that it holds also
for smoothed Sinkhorn and MMD, we leave the proofs for future works. Based on the property in Proposition~\ref{proposition:2-level-noise}, we show some specific properties of the metric 
with respect to the noise level $\sigma$. 
\begin{proposition}
\label{prop:decresaing_sgswd}
 $\gssd^p(\mu,\nu)$ is decreasing with respect to 
    $\sigma$ and we have $\lim_{\sigma \rightarrow 0}  \gssd^p(\mu,\nu) = \divg^p(\mu,\nu).$
\end{proposition}
The proof of Proposition~\ref{prop:decresaing_sgswd} comes straightforwardly from Proposition~\ref{proposition:2-level-noise} by taking $\sigma_2 = \sigma$   and letting $\sigma_1 \rightarrow 0$. 
This property interestingly states that the $\gssd^p$ recovers the sliced divergence when the noise level vanishes. 
{We end up this section by providing a relation between Gaussian-smoothed sliced Wasserstein distances under two noise levels.

\begin{proposition}
\label{proposition:GS-SWD_sigma_1_2}
Let $0\leq \sigma_1\leq \sigma_2$ be two noise levels. Then, one has
\begin{align*}
\gssww^p(\mu,\nu) &\leq 2^{p-1} \gsswww^p(\mu,\nu)+ 2^{\frac{5p}{2}} (\sigma_2^2 - \sigma_1^2)^{p}.
\end{align*}
\end{proposition}

\subsubsection{Continuity}
Now we analyze the continuity properties of some $\gssd^p(\mu,\nu)$ w.r.t. the noise level.

\begin{proposition}
\label{prop:continuity_of_gs_swd}
 For any two distributions $\mu$ and $\nu$ for which
	the sliced Wasserstein is well-defined,
the Gaussian-smoothed sliced Wasserstein distance is continuous w.r.t. to $\sigma$.
\end{proposition}

\begin{proposition}
\label{prop:continuity_smoothed_mmd}
{Assume that the kernel defining the maximum mean discrepancy $(\MMD)$ divergence is bounded. Then the Gaussian-smoothed sliced $\gssmmd$ is continuous w.r.t. to $\sigma$.}
\end{proposition}

The above propositions show that most distribution divergences are continuous with respect to $\sigma$ under mild conditions.


\begin{figure}
	\centering
	\includegraphics[width=5cm]{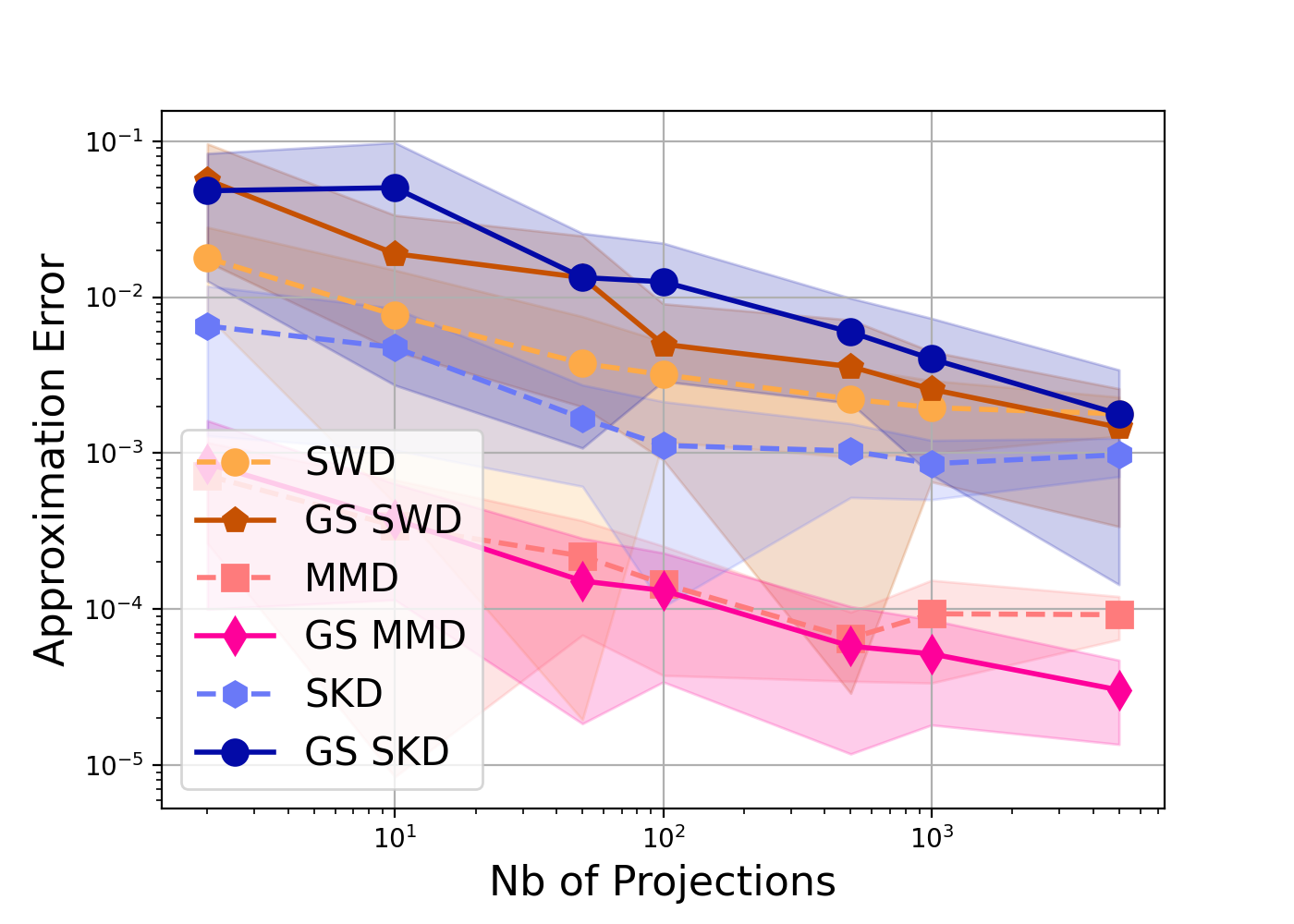}   
	\caption{Absolute difference between the approximated Monte Carlo approximation
		of all divergences compared to the true one (evaluated with $10,000$ number of projections).
		The  two sets of $500$ samples in $\R^{50}$ are  randomly drawn from $\mathcal{N}(0,\mathbf{I})$.
		The Gaussian-smoothed sliced divergences are parameterized with $\sigma=3$.
		\label{fig:approx}}  
\end{figure}
    \section{Numerical Experiments}  \label{sec:expe}
    \label{sec:expe}
    In this section, we report on a series of experiments that support the  established theoretical results.
        We also highlight the usefulness of the findings     in a context of privacy-preserving domain adaptation problem.

\subsection{Supporting the theoretical results}

\paragraph{Sample complexity.}
The first  experiment (see Figure \ref{fig:divsamples}) analyzes the sample complexity of different base divergences. 
It shows that the sample complexity stays similar to the one of their original 
and sliced counterparts up to a constant (see Proposition~\ref{theorem:sample_double_empirical_complexity}). 
For this purpose, we have considered samples in $\mathbb{R}^d$ randomly drawn from a Normal distribution 
$\mathcal{N}(0,\mathbf{I})$.   For the Sinkhorn divergence, the entropy regularization has been set
to $0.1$ and for MMD, we used a Gaussian kernel for which the bandwidth has been set
to the mean of all pairwise distances between samples. 
The number of projections has been fixed
to $L=50$ and we perform 20 runs per experiment. For the first study, the convergence rate has been evaluated by increasing the samples number
up to 25,000 with fixed dimension $d=50$. For the second one, we vary both the dimension and the number of samples.

Figure \ref{fig:divsamples} shows the sample complexity of some sliced divergences, respectively noted as SWD, SKD and MMD for Sliced Wasserstein distance, Sinkhorn divergence and Maximum Mean discrepancy 
and their Gaussian-smoothed sliced versions, named as GS SWD, GS SKD and GS MMD. On the top plot, we can see that all 
Gaussian-smoothed sliced divergences preserve the complexity rate with just a slight to moderate overhead. The worst
difference is for Sinkhorn divergence, while MMD almost comes for free
in term of complexity. From the bottom plot where sample complexities for different dimensions $d$
are given, we confirm the finding that  Gaussian smoothing keeps the independence of the convergence rate to the dimension of sliced divergences.  

Two other experiments on the sample complexity and identity of indiscernibles are also reported in the supplementary material.

\paragraph{Projection complexity.}
We have also investigated the impact of the number of projections when estimating the distance
between two sets of $500$ samples drawn from the same distribution, $\mathcal{N}(0,\mathbf{I})$.
Figure  \ref{fig:approx} plots the approximation error  between the true expectation of the
sliced divergences (computed for a number of $L=10,000$ projections) and its approximated
versions. We remark that, for all methods, the error ranges within $10$-fold when approximating
with $50$ projections and decreases with the number of projections. 

\paragraph{Performance path on the impact of the noise parameter.} Since the Gaussian smoothing parameter $\sigma$
is  key in a privacy preserving context, as it impacts on the level of privacy of the Gaussian
mechanism, we have analyzed its impact on the  smoothed sliced divergence.
We have reproduced the experiment for the sample complexity but with different values
of $\sigma$. The number of projections has been set to $50$.  
Figure \ref{fig:samplenoise} shows these sample complexities.
The first very interesting
point to note is that the smoothing parameter has almost no effect on the GS MMD sample
complexity. For the GS SWD and GS SKD divergences, instead, the smoothing
tends to increase the divergence at fixed number of samples. Another interpretation
is that to achieve a given value of divergence, one needs more far samples when the
smoothing is larger (\emph{i.e.} for getting a given divergence value at $\sigma=5$, one needs almost
$10$-fold more samples for $\sigma=15$). 
This overhead of samples needed when smoothing increases is properly
described, for the Gaussian-smoothed sliced SWD in our Proposition~\ref{theorem:sample_double_empirical_complexity}, as the sample complexity depends
on the moments of the Gaussian.

\begin{figure}
	\centering
    \includegraphics[width=5cm]{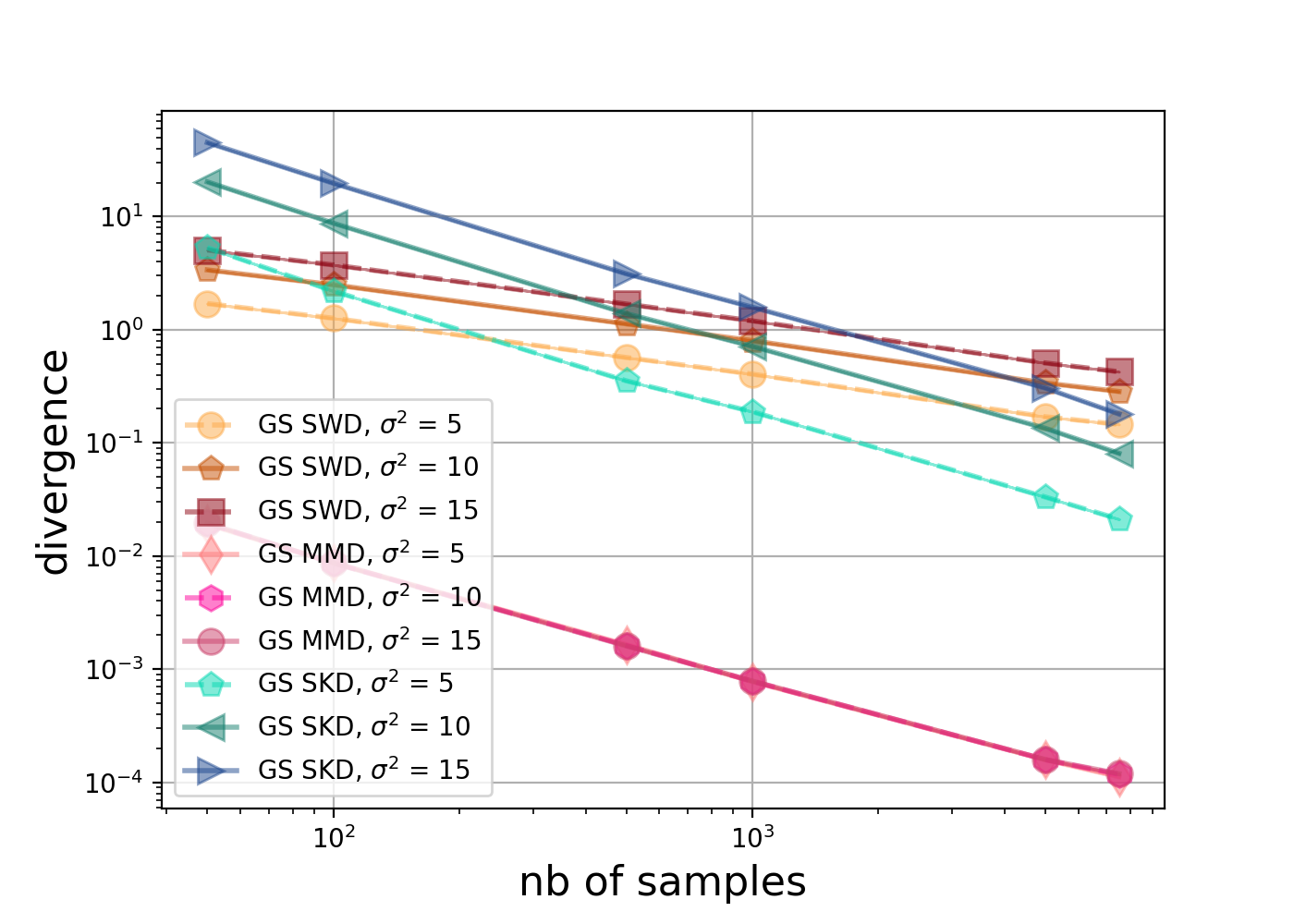}   
    \caption{Measuring the divergence between two sets of samples in $\R^{50}$ drawn
    from   $\mathcal{N}(0,\mathbf{I})$. We plot the sample complexity for different Gaussian-smoothed sliced 
    divergence at different level of noises.
    \label{fig:samplenoise}}
\end{figure}

As for conclusion from these analyses, we highlight that the Gaussian-smoothed sliced MMD seems
to present several strong benefits: its sample complexity does not depend on 
the dimension and seems to be the best one among the
divergence we considered. More interestingly, it is not impacted by the amount
of Gaussian smoothing and thus not impacted by a desired privacy level.

\subsection{Domain adaptation with $\gssw$}\begin{figure*}[t]
	\begin{center}
		\includegraphics[width=5cm]{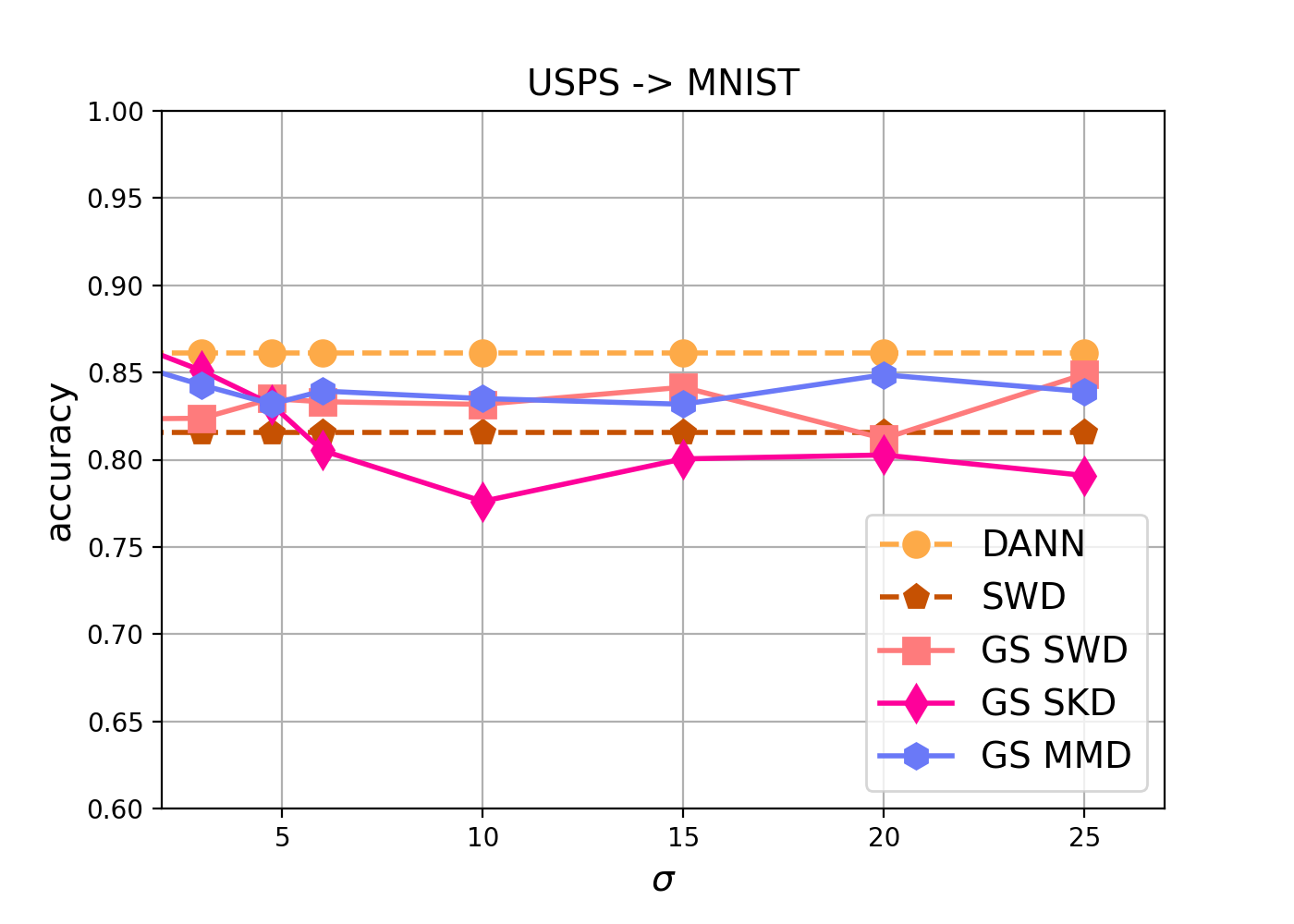}
		\includegraphics[width=5cm]{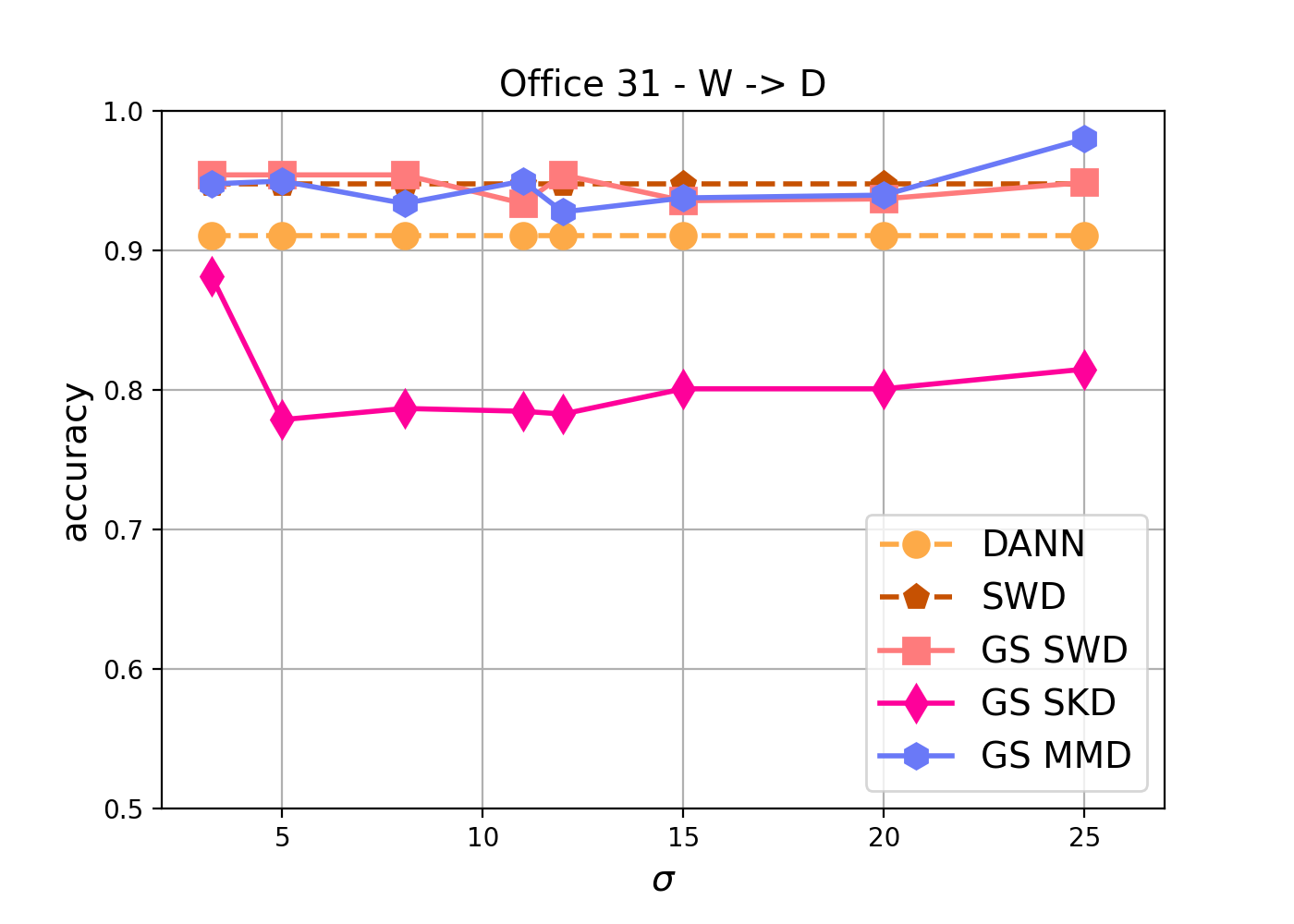}
		\includegraphics[width=5cm]{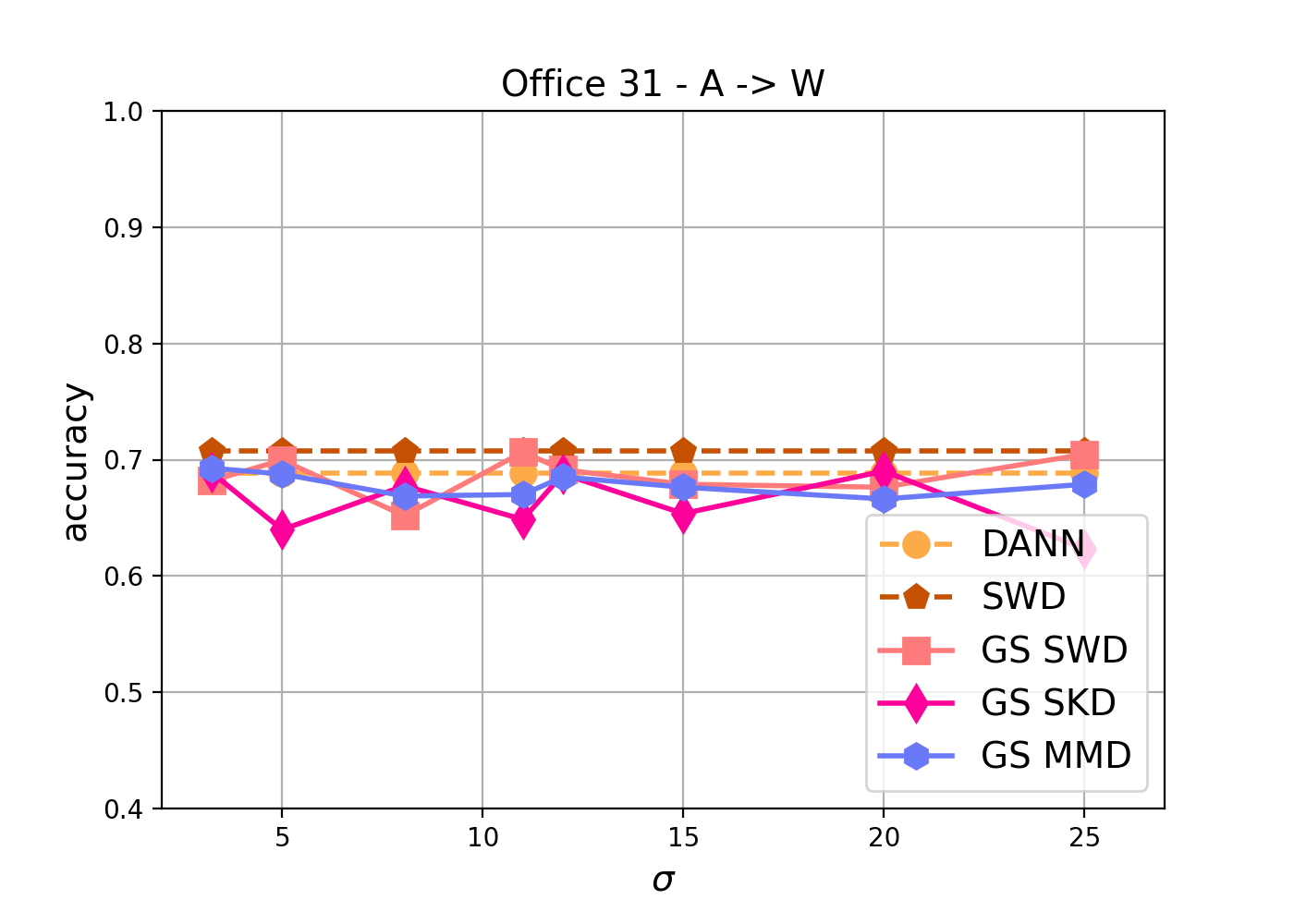}
	\end{center}
		\caption{Domain adaptation performances using different divergences on distributions with respect to the Gaussian smoothing. (Left) USPS to MNIST. (Middle) Office-31 Webcam to DSLR. (Right) Office-31 Amazon to Webcam. \label{fig:da} }
\end{figure*}

\begin{figure*}[htbp]
	\begin{center}
		\includegraphics[width=5cm]{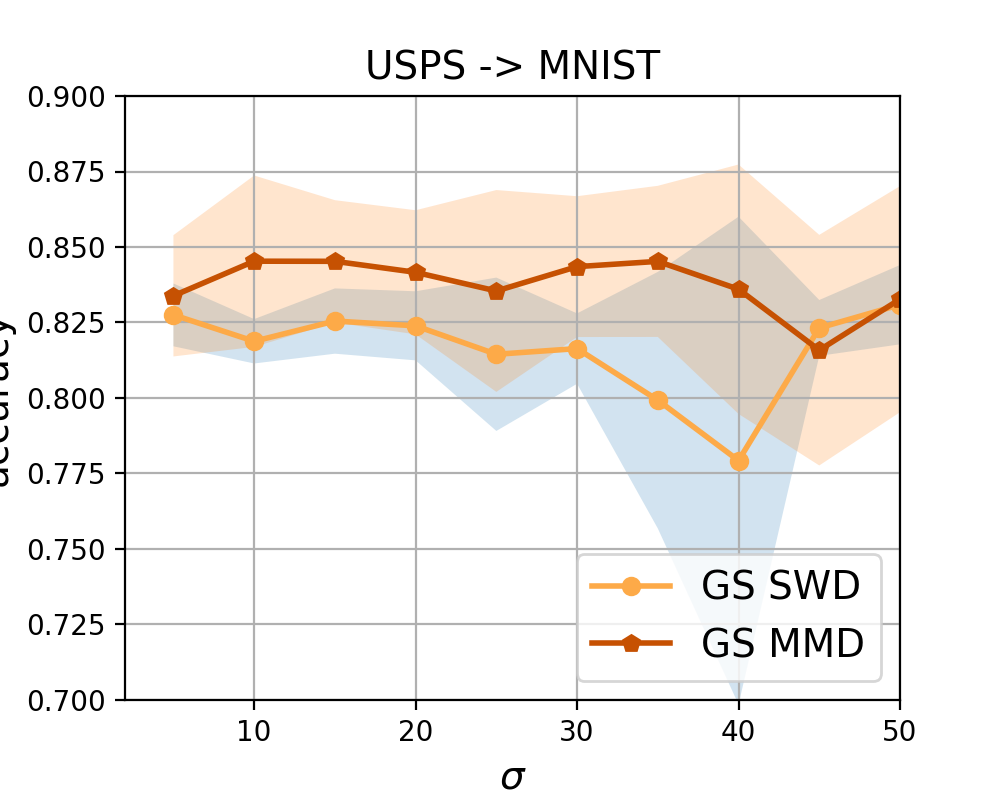}
		\includegraphics[width=5cm]{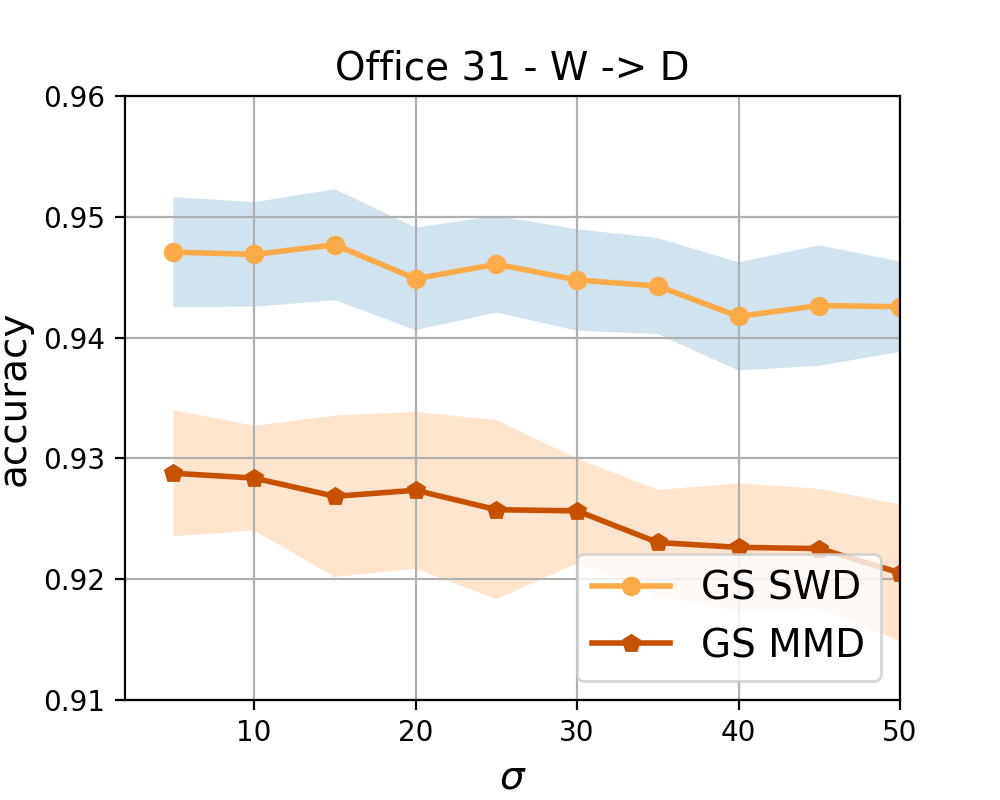}
		\includegraphics[width=5cm]{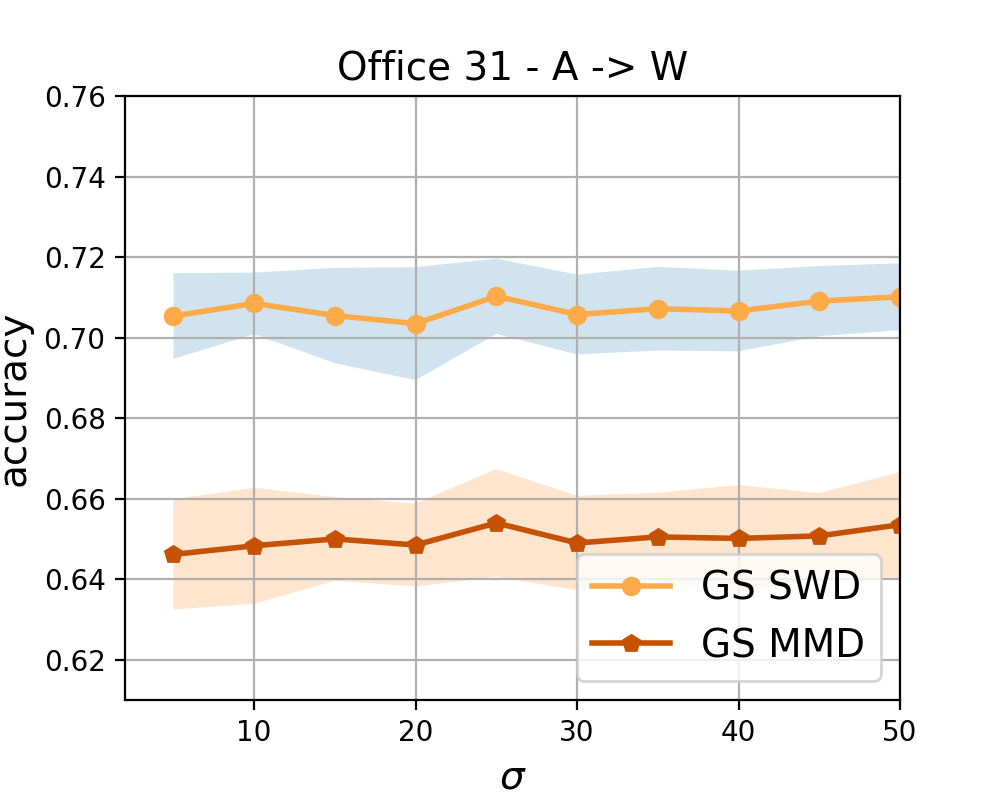}
	\end{center}
	\caption{
		Domain adaptation performances using different divergences on distributions with respect to the Gaussian smoothing using \textbf{one-epoch-fine-tuned} models. (Left) USPS to MNIST. (Middle) Office-31 Webcam to DSLR. (Right) Office-31 Amazon to Webcam. \label{fig:daref} }
\end{figure*}

As an application, we have considered the problem of unsupervised domain adaptation for a classification task. In this
context, given source examples $\X_s$ and their label $\y_s$ and unlabeled
	target examples $\X_t$, our goal is to design a classifier $h(\cdot)$ learned from 
	the source examples that generalizes well on the target ones. A classical
    approach consists in learning a representation mapping  $g(\cdot)$ that leads to
     invariant latent representations, invariance being measured as a distance
      between empirical distributions of mapped source and target samples. 
      Formally, this leads to the following  problem
     \begin{equation*}
     	\label{eq:domain_adapt}
\min_{g,h}\big\{\mathcal{L}_c(h(g(\X_s)),\y_s) + \mathcal{D}(g(\X_s), g(\X_t))\big\}
     \end{equation*}
      where $\mathcal{L}_c$ can be the cross-entropy loss or a quadratic loss and 
    $\mathcal{D}$ a divergence between empirical distributions,
   in our case, $\mathcal{D}$ will be any Gaussian-smoothed sliced divergence.
     We solve this problem through stochastic gradient descent, 
     similarly to many approaches that use sliced Wasserstein distance as a 
     distribution distance~\cite{lee2019sliced}.
Note that, in practice,  using a smoothed divergence preserves the privacy of the 
target samples as shown by~\citep{pmlr-v139-rakotomamonjy21a}.

			When performing such model adaptation, a privacy/utility trade-off that has to be handled. In practice, one would prefer the most private model while not hurting its performance. Hence, one would seek the largest noise level $\sigma>0$ to use  while preserving accuracy on target domain.
		Hence, it is useful to evaluate how the model performs  on a range
	of noise level (hence, privacy level). This can be computationally expensive
	at it requires to fully train several models on hundreds of epochs. 
		Instead, we leverage on the continuity of our $\gssd$ to
	 employ a fine-tuning strategy:
		we train a domain adaptation model for the largest desired value of $\sigma$ (over the full number of epochs) and when $\sigma$ is decreased, we just fine-tune the
	lasted model by training on only one epoch. 

Our experiments evaluate the studied  Gaussian-smoothed sliced divergences in classical unsupervised domain adaptation.
 We have considered two datasets: 
a handwritten digit recognition (USPS/MNIST) and Office 31 datasets.

In our first analysis, we have compared our {$\gssd$ performances}  with non-smoothed divergences. The first one is the 
  sliced Wasserstein distance (SWD) \cite{lee2019sliced} and the second one is the Jenssen-Shannon approximation
  based on adversarial approach, known as DANN \cite{ganin2015unsupervised}. 
  For all methods and for each dataset, we used
the same neural network architecture for representation mapping and for
classification. Approaches differ only on how distance between distributions have been computed. Here for each noise value $\sigma$, we have trained the model from scratch for $100$ epochs.
Results are depicted in Figure \ref{fig:da}. For the two problems, we can see that performances obtained with the Gaussian-smoothed sliced Wasserstein or MMD divergences are similar to those obtained with DANN or SWD across all ranges of noise. 
The smoothed version of Sinkhorn is less stable and induces
a slight loss of performance. Owing to the metric property and
the induced weak topology, the privacy preservation comes almost
without loss of performance  in this domain adaptation context.  

{In the second analysis, we have studied the privacy/utility trade-off  when fine-tuning models, using only one epoch, for decreasing values of $\sigma$. Results are shown in Figure~\ref{fig:daref}. They highlight	that
depending on the data and the used smoothed divergence, performance varies between  one percent for Office 31 to four percent for USPS to MNIST. Note that except for the largest value of $\sigma$,  we are training a model using only one epoch instead of a hundred. A very large gain in complexity is thus achieved for swiping the full range of noise level.
 Hence depending on the importance this slight drop in performance will have, it is worth using a large value of $\sigma$ and preserving strong privacy or go through a validation procedure of several (cheaply obtained) models.}


\section{Conclusion}
\label{sec:conclu}
This work provided the properties of Gaussian-smoothed sliced divergences for comparing distributions. 
We derived several theoretical results related to their topological and statistical properties and showed, under mild conditions on their base divergences, the smoothing and slicing operations preserves the metric property. 
From a statistical point of view, we introduced the double empirical distribution and focused on the sample complexity of the smoothed sliced Wasserstein distance and we proved that it converges with a rate $O(n^{-1/2}).$  
We furhter analyzed the behavior of these divergences on domain adaptation problems and confirm the fact that using those divergences yields only to slight loss of performances while preserving privacy.
An important direction for future research is establishing order statistic representations for Sinkhorn and MMD divergences. 
Furthermore, in the obtained bound we use upper bound of higher moments of the smoothing distribution, that would think to consider non Gaussian smoothing distribution enjoying this property.





\clearpage
\appendix

\appendix

\section{Proofs and additional theoretical results}
\label{sec:appendix_proofs}

In the following sections, we give the proofs of the theoretical guarantees given in the main of the paper. For a sake of completeness, we add  other results that we consider interesting. 

\subsection{Proof of Theorem~\ref{theorem:proof_topology}: $\gssd$ is a proper metric on $\mathcal{P}_p(\R^d) \times \mathcal{P}_p(\R^d)$} \label{sub:proof_of_theorem_theorem:proof_topology}

Before starting the proof, we add this notation: the characteristic function of a probability distribution
$\mu \in \cP(\R^d)$  is $\varphi_\mu(t) = \E_\mu[e^{iX^\top t}]$. Given this definition,
similarly to the Fourier transform, the characteristic function of the convolution of
two probability distributions readsas  $\varphi_{\nu * \mu}(t) =\varphi_{\nu}(t)\cdot\varphi_{\mu}(t)$. 

$\bullet$ {\it Non-negativity (or symmetry).} The non-negativity (or symmetry) follows directly from the non-negativity (or symmetry) of $\divg^p$, see Definition \ref{def:Gaussian_smoothed_sliced_Div}.\\
$\bullet$ {\it Identity  property.} If the base divergence $\divg^p$ satisfies the identity property in one dimensional measures, then for any $\mu \in \mathcal{P}_p(\R^d)$ and $\u \in \mathbb{S}^{d-1}$, one has that $\divg^p(\mathcal{R}_\u \mu * \mathcal{N}_\sigma,\mathcal{R}_\u \mu* \mathcal{N}_\sigma) =0,$ hence, by Definition \ref{def:Gaussian_smoothed_sliced_Div}, $\gssd^p(\mu,\mu) = 0.$
Let us now prove the fact that for any $\mu, \nu \in \mathcal{P}_p(\R^d), \gssd^p(\mu,\nu) =0$ entails $\mu = \nu$ a.s.
On one hand, $\gssd^p(\mu,\nu) = 0$ gives the fact that $\divg^p(\mathcal{R}_\u \mu * \mathcal{N}_\sigma,\mathcal{R}_\u \nu* \mathcal{N}_\sigma) = 0$ for $u_d$-almost every $\u \in \mathbb{S}^{d-1},$ hence $\mathcal{R}_\u \mu * \mathcal{N}_\sigma = \mathcal{R}_\u \nu* \mathcal{N}_\sigma$ for $u_d$-almost every $\u \in \mathbb{S}^{d-1}.$ Following the techniques in proof of Proposition 5.1.2 in~\cite{bonnotte:tel-00946781}, for any measure $\eta \in \mathcal{P}(\R^m)$ (with $m\geq 1$), $\mathcal{F}[\eta](\cdot)$ stands for the Fourier transform of {$\eta$} and is given as {$\mathcal{F}[\eta](\v) = \int_{\R^m} e^{-i\s^\top \v}\diff\eta(\s)$ for any $\v \in \R^m.$}
Then 
\begin{align*}
\mathcal{F}[\mathcal{R}_\u \mu * \mathcal{N}_\sigma](v) &= \int_\R e^{-i v {t}} \diff(\mathcal{R}_\u \mu * \mathcal{N}_\sigma)(t)\\
&=\int_{\R}\int_{\R}e^{-i(r+t)v}\diff\mathcal{R}_\u\mu(r) \diff\mathcal{N}_\sigma(t) \quad ({\text{by the definition of the convolution operator}})\\
&= \int_{\R^d}\int_{\R} e^{-i(\langle \u, {\s}\rangle+t)v}
\diff\mu(\s) \diff\mathcal{N}_\sigma(t) \quad ({\text{by the definition of Radon Transform}})\\
&= \int_{\R} e^{-itv}\diff\mathcal{N}_\sigma(t) \int_{\R^d} e^{-i(\langle \u, s\rangle)v}
\diff\mu(\s)\\
&= \mathcal{F}[\mathcal{N}_\sigma](v)\mathcal{F}[\mu](v\u).
\end{align*}
Since for $u_d$-almost every $\u\in \mathbb{S}^{d-1}, \mathcal{R}_\u \mu * \mathcal{N}_\sigma = \mathcal{R}_\u \nu* \mathcal{N}_\sigma$, and hence $\mathcal{F}[\mathcal{R}_\u \mu * \mathcal{N}_\sigma] = \mathcal{F}[\mathcal{R}_\u \nu* \mathcal{N}_\sigma]
\Leftrightarrow \mathcal{F}[\mathcal{N}_\sigma]\mathcal{F}[\mu] = \mathcal{F}[\mathcal{N}_\sigma]\mathcal{F}[\nu] \,\, ({\text{by the Fourier transform of the convolution}}) \,\, \Leftrightarrow \mathcal{F}[\mu] = \mathcal{F}[\nu]$. Since the Fourier transform is injective, we conclude that $\mu=\nu.$\\
$\bullet${\it Triangle inequality.} Assume that $\divg$ is a metric and let $\mu, \nu, \eta \in \mathcal{P}_p(\R^d).$ We then have  
\begin{align*}
\gssd(\mu,\nu)
& = \Big\{\int_{\mathbb{S}^{d-1}}\divg^p(\mathcal{R}_\u\mu * \mathcal{N}_\sigma,\mathcal{R}_\u \nu* \mathcal{N}_\sigma) u_d(\u)\diff \u\Big\}^{1/p}\\
&\leq \Big\{\int_{\mathbb{S}^{d-1}}\Big(\divg(\mathcal{R}_\u\mu * \mathcal{N}_\sigma,\mathcal{R}_\u \eta* \mathcal{N}_\sigma) + \divg(\mathcal{R}_\u\eta * \mathcal{N}_\sigma,\mathcal{R}_\u \nu* \mathcal{N}_\sigma)\Big)^pu_d(\u)\diff \u\Big\}^{1/p}\\
&\underbrace{\leq}_{(\star)} \Big\{\int_{\mathbb{S}^{d-1}}\Big(\divg^p(\mathcal{R}_\u\mu * \mathcal{N}_\sigma,\mathcal{R}_\u \eta* \mathcal{N}_\sigma)u_d(\u)\diff \u\Big\}^{1/p}\\
& \qquad + \Big\{\int_{\mathbb{S}^{d-1}} \divg^p(\mathcal{R}_\u\eta * \mathcal{N}_\sigma,\mathcal{R}_\u \nu* \mathcal{N}_\sigma)\Big)^pu_d(\u)\diff \u\Big\}^{1/p}\\
&= \gssd(\mu,\eta) + \gssd(\eta,\nu),
\end{align*}
where inequality in $(\star)$ follows from the application of Minkowski inequality.

\subsection{Proof of Theorem~\ref{theorem-weaktopo}: $\gssd$ metrizes the weak topology}
The proof is done by double implications:

``$\Rightarrow$'' Assume that $\mu_k \Rightarrow\mu$. 
Fix $\u \in \mathbb{S}^{d-1},$ the mapping $\u \mapsto \mathcal{R}_\u$ is continuous from $\R^d$ to $\R$, then an application of continuous mapping theorem~\citep{athreya2006measure} entails that $\mathcal{R}_\u\mu_k \Rightarrow\mathcal{R}_\u\mu$. By Lévy's continuity theorem~\citep{athreya2006measure}  $\mathcal{R}_\u\mu_k* \mathcal{N}_\sigma\Rightarrow\mathcal{R}_\u\mu* \mathcal{N}_\sigma$. Therefore, $\lim_{k\rightarrow \infty} \divg(\mathcal{R}_\u\mu_k, \mathcal{R}_\u\mu* \mathcal{N}_\sigma) = 0.$ Since we suppose that the divergence $\divg$ is bounded, then there exists $K\geq 0$ such that for any $k$, $\divg^p(\mathcal{R}_\u\mu_k, \mathcal{R}_\u\mu* \mathcal{N}_\sigma) \leq K.$ An application of bounded convergence theorem  yields
\begin{equation*}
\lim_{k\rightarrow\infty} \gssd^p(\mu_k,\mu) = \lim_{k\rightarrow\infty}\int_{\mathbb{S}^{d-1}} \divg^p(\mathcal{R}_\u\mu_k * \mathcal{N}_\sigma,\mathcal{R}_\u \mu* \mathcal{N}_\sigma)u_d(\u)\diff \u = 0.
\end{equation*}
``$\Leftarrow$'' (By contrapositive). Suppose that $\mu_k$ doesn't converge weakly to $\mu$ and assume that $\lim_{k\rightarrow\infty} \gssd^p(\mu_k,\mu)=0$. On one hand, since $\R^d$ is a complete separable space then  the weak convergence is equivalent to the convergence corresponding to  Lévy-Prokhorov distance $\Lambda$  defined as:
The  Lévy-Prokhorov
distance $\Lambda(\eta, \zeta)$ between $\eta, \zeta \in \mathscr{P}((E,\rho), \mathcal{T})$ (space of probability measures on a measurable metric space)  is given by:
\begin{equation*}
\Lambda(\eta, \zeta) = \inf\limits_{\varepsilon >0} \{\eta(A) < \zeta(A^\varepsilon) + \varepsilon, \quad \zeta(A) < \eta(A^\varepsilon) + \varepsilon, \quad \text{for all } A \in \mathcal{T}\}, 
\end{equation*}
\text{ where }$A^\varepsilon =\{x \in E: \rho(x, A) < \varepsilon\}.$
Hence there exists $\varepsilon >0$ and a subsequence $\{\mu_{s(k)}\}_{k\in \mathbb N}$ such that 
$\Lambda(\mu_{s(k)}, \mu) > \varepsilon.$ One the other hand, we have $\lim_{k\rightarrow\infty} \gssd^p(\mu_{s(k)},\mu)=0$, that 
is equivalent to $\{\divg(\mathcal{R}_\u\mu_{s(k)} * \mathcal{N}_\sigma,\mathcal{R}_\u \nu* \mathcal{N}_\sigma)\}_k$ converges to $0$ in $L^p(\mathbb{S}^{d-1}) = \{f:\mathbb{S}^{d-1} \rightarrow \R| \int_{\mathbb{S}^{d-1}}f(\u) u_d(\u) \diff u < \infty\}.$ Since the $L^p$-convergence entails the point-wise convergence~\citep{khoshnevisanprobability}, there exists a subsequence 
$\{\mu_{s(t(k))}\}_k$ such that $\lim\limits_{k \rightarrow \infty}\divg(\mathcal{R}_\u\mu_{s(t(k))} * \mathcal{N}_\sigma,\mathcal{R}_\u \mu* \mathcal{N}_\sigma) = 0$ almost {everywhere} for all $\u \in \mathbb{S}^{d-1}.$ Recall that the divergence $\divg$ metrizes the weak convergence in $\mathcal{P}(\R)$
then $\mathcal{R}_\u\mu_{s(t(k))} * \mathcal{N}_\sigma\Rightarrow \mathcal{R}_\u \mu* \mathcal{N}_\sigma$ almost {everywhere} for all $\u \in \mathbb{S}^{d-1}.$ 
Therefore, $\mathcal{R}_\u\mu_{s(t(k))}\Rightarrow \mathcal{R}_\u \mu$ almost {everywhere} for all $\u \in \mathbb{S}^{d-1}.$ Using Cramér-Wold device~\citep{Huber2011}, we get $\mu_{s(t(k))}\Rightarrow \mu.$ Since the  Lévy-Prokhorov distance metrizes the weak convergence, it entails that $\lim\limits_{k \rightarrow \infty} \Lambda(\mu_{s(t(k))}, \mu_k) = 0$, that contradicts the fact that $\Lambda(\mu_{s(k)}, \mu) > \varepsilon.$ We then conclude by contrapositive that $\mu_k \Rightarrow \mu.$

\subsection{Proof of Proposition~\ref{prop_lower_semicont}: $\gssd$ is lower semi-continuous} 
Recall that the base divergence $\divg$ is lower semi-continuous w.r.t.  the weak topology in $\mathcal{P}({\R})$, namely for every sequence of measures $\{\mu'_k\}_{k\in \mathbb N}$ and $\{\nu'_k\}_{k\in \mathbb N}$ in $\mathcal{P}({\R})$  such that $\mu'_k \Rightarrow \mu'$ and $\nu'_k \Rightarrow \nu'$, one has $\divg(\mu', \nu') \leq \liminf\limits_{k \rightarrow \infty}\divg(\mu'_k, \nu'_k)$.
\\
Now, let $\{\mu_k\}_{k\in \mathbb N}$ and $\{\nu_k\}_{k\in \mathbb N}$ are two sequences of measure in $\mathcal{P}_p(\R^d)$ such that $\mu_k \Rightarrow \mu$ and $\nu_k \Rightarrow \nu$. By continuous mapping theorem~\citep{bowers2014introductory} and Levy's continuity theorem, we obtain $\mathcal{R}_\u \mu_k* \mathcal{N}_\sigma\Rightarrow \mathcal{R}_\u \mu* \mathcal{N}_\sigma$ and $\mathcal{R}_\u \nu_k* \mathcal{N}_\sigma\Rightarrow \mathcal{R}_\u \nu* \mathcal{N}_\sigma$ for all $\u \in \mathbb{S}^{d-1}.$
Since the base divergence $\divg$ is a lower semi-continuous with respect to weak topology in $\mathcal{P}(\R)$, then
\begin{equation*}
\divg^p(\mathcal{R}_\u \mu* \mathcal{N}_\sigma, \mathcal{R}_\u \nu* \mathcal{N}_\sigma) \leq \big(\liminf\limits_{k \rightarrow \infty} \divg(\mathcal{R}_\u \mu_k* \mathcal{N}_\sigma, \mathcal{R}_\u \nu_k* \mathcal{N}_\sigma)\big)^p \leq \liminf\limits_{k \rightarrow \infty} \divg^p(\mathcal{R}_\u \mu_k* \mathcal{N}_\sigma, \mathcal{R}_\u \nu_k* \mathcal{N}_\sigma).
\end{equation*}
It gives
\begin{equation*}\gssd^p(\mu,\nu) \leq \int_{\mathbb{S}^{d-1}}\liminf\limits_{k \rightarrow \infty} \divg^p(\mathcal{R}_\u \mu_k* \mathcal{N}_\sigma, \mathcal{R}_\u \nu_k* \mathcal{N}_\sigma) u_d(\u) \diff\u.
\end{equation*}
Furthermore, by application of Fatou's lemma~\citep{bowers2014introductory}, we get
\begin{equation*}
\gssd^p(\mu,\nu) \leq\liminf\limits_{k \rightarrow \infty}\int_{\mathbb{S}^{d-1}} \divg^p(\mathcal{R}_\u \mu_k* \mathcal{N}_\sigma, \mathcal{R}_\u \nu_k* \mathcal{N}_\sigma) u_d(\u) \diff\u = \liminf\limits_{k \rightarrow \infty} \gssd^p(\mu_k,\nu_k),
\end{equation*}
which is the desired result.

\subsection{Proofs of statistical properties}

\subsubsection{Proof of Lemma~\ref{lem:law_of_sum_of_smooth_projections}}

Straighforwardly, for every Borelian $I \in \mathcal{B}(\R)$, we have 
\begin{align*}
\mathcal{R}_\u \hat\mu_n * \mathcal{N}_\sigma (I) &= \int_r\int_s\mathbf{1}_I(r+s) \diff\{\frac1n\sum_{i=1}^n \delta_{\u^\top X_i}\}(r)\diff\mathcal{N}_\sigma(s)\\
&=  \frac 1n \sum_{i=1}^n \int_s\mathbf{1}_I(\u^\top X_i+s)f_{\mathcal{N}_\sigma}(s)\diff s\\
&=  \frac 1n \sum_{i=1}^n \int_{s'}\mathbf{1}_I({s'})f_{\mathcal{N}_\sigma}(s' - \u^\top X_i)\diff s'\\
&= \frac 1n \sum_{i=1}^n \int_{s'}\mathbf{1}_I({s'})f_{\mathcal{N}(\u^\top X_i, \sigma^2)}(s')\diff s' \quad ({\text{since } \tiny{f_{\mathcal{N}_\sigma}(s'-\u^\top X_i) 
= f_{\mathcal{N}(\u^\top X_i, \sigma^2)}(s')}})\\
&= \frac 1n \sum_{i=1}^n{\mathcal{N}(\u^\top X_i, \sigma^2)}(I).
\end{align*}
Thanks to Theorem of Cramér and Wold~\citep{cramerwold1936}, we conclude the equality between the measures $\mathcal{R}_\u \hat\mu_n * \mathcal{N}_\sigma = \frac 1n \sum_{i=1}^n{\mathcal{N}(\u^\top X_i, \sigma^2)}.$

\subsubsection{Proof of Proposition~\ref{theorem:sample_double_empirical_complexity}}  

For this proof and the following ones, we use frequently the triangle inequality for Wasserstein distances between the quantities $\hat{\hat\mu}_{n},$ $\frac 1n \mathcal{N}_\sigma(u^\top X_i, \sigma^2)$ and  $\mathcal{R}_\u \mu* \mathcal{N}_\sigma.$ 
 On one hand, using triangle inequality of Wasserstein distance, we have 
\begin{align*}
\E_{\mu^{\otimes_n}| \mathcal{N}_\sigma^{\otimes_n}}[{\hatgssw^p}({\hat\mu}_n,\mu)] &=  \int_{\mathbb{S}^{d-1}} \E_{\mu^{\otimes_n}| \mathcal{N}_\sigma^{\otimes_n}}[\wass^p_p(\hat{\hat\mu}_{n}, R_\u\mu * \mathcal{N}_\sigma )]u_d(\u)\diff\u \leq {\bf I} + {\bf II}
\end{align*}
where 
\begin{align*}
{\bf I} = 2^{p-1} \int_{\mathbb{S}^{d-1}} \E_{\mu^{\otimes_n}| \mathcal{N}_\sigma^{\otimes_n}}\Big[\wass^p_p\Big(\hat{\hat\mu}_{n}, \frac 1n \sum_{i=1}^n \mathcal{N}(\u^\top X_i, \sigma^2)\Big)\Big]u_d(\u)\diff\u 
\end{align*}
and 
\begin{align*}
{\bf II} = 2^{p-1} \int_{\mathbb{S}^{d-1}} \E_{\mu^{\otimes_n}| \mathcal{N}_\sigma^{\otimes_n}}\big[\wass^p_p\big( \frac 1n \sum_{i=1}^n \mathcal{N}(\u^\top X_i, \sigma^2), R_\u\mu * \mathcal{N}_\sigma )\big)\big]u_d(\u)\diff\u 
\end{align*}
The proof is based on two steps to control the quantities ${\bf I}$ and ${\bf II}$.\\
\noindent {\it \underline{Step 1: Control of} ${\bf I}$.}\\
Let us state the following lemma:
\begin{lemma}[See proof of Theorem 1 in~\cite{fournier2015}]
\label{lemma:thm1-fournier}

Let $\eta \in \mathcal{P}(\R)$ and let $p\geq 1$. Assume that $M_q(\eta)<\infty$ for some $q>p.$ There exists a constant $C_{p,q}$ depending only on $p,q$ such that, for all $n\geq 1,$
\begin{align*}
\E[\wass^p_p(\hat\eta_n, \eta)] \leq C_{p,q}M_q(\eta)^{p/q} \Delta_{n}(p,q),
\end{align*}
where 
\begin{equation*}
\Delta_{n}(p,q) = 
\begin{cases}
n^{-1/2}{\bf{1}}_{q >2p }, \\
n^{-1/2}\log(n){\bf{1}}_{q=2p}\\
n^{-(q-p)/q}{\bf{1}}_{p<q<2p}.\end{cases}.
\end{equation*}
\end{lemma}
We note that $\hat{\hat\mu}_{n}$ is an empirical version of the Gausian mixture $\frac 1n \sum_{i=1}^n \mathcal{N}_\sigma(u^\top X_i, \sigma^2)$. Then, by application of Lemma~\ref{lemma:thm1-fournier}, we get 
\begin{align*}
\E_{\mu^{\otimes_n}| \mathcal{N}_\sigma^{\otimes_n}}\big[\wass^p_p\big(\hat{\hat\mu}_{n}, \frac 1n \sum_{i=1}^n \mathcal{N}(\u^\top X_i, \sigma^2)\big)\big] \leq C_{p,q} \E_{\mu^{\otimes_n}}\Big[M_q^{p/q}\Big(\frac 1n \sum_{i=1}^n \mathcal{N}(\u^\top X_i, \sigma^2)\Big)\Big] \Delta_n(p, q).
\end{align*}
Let us first upper bound the $q$-th moment of $M_q\Big(\frac 1n \sum_{i=1}^n \mathcal{N}(\u^\top X_i, \sigma^2)\Big)$, for all $q\geq 1.$
For all $\u\in \mathbb{S}^{d-1} $, we have 
\begin{align*}
M_q\Big(\frac 1n \sum_{i=1}^n \mathcal{N}(\u^\top X_i, \sigma^2)\Big) 
&= \int_{\R}|t|^q \text{d}(\frac 1n \sum_{i=1}^n \mathcal{N}(\u^\top X_i, \sigma^2))(t) =  \frac 1n \sum_{i=1}^n M_q(|Z_{i,\u}|^q),
\end{align*}
where $Z_{i,\u}\sim \mathcal{N}(\u^\top X_i, \sigma^2)).$
By Equation (17) in~\cite{winkelbauer2014moments}
we have 
\begin{align*}
M_q\Big(\frac 1n \sum_{i=1}^n \mathcal{N}(\u^\top X_i, \sigma^2)\Big) = \frac 1n \frac{2^{q /2}\sigma^q}{\sqrt{\pi}}\Gamma(\frac{q +1}{2})\sum_{i=1}^n  {}_1 F_1\big(-\frac q 2, \frac 12; \frac{-(\u^\top X_i)^2}{2\sigma^2}\big).
\end{align*}
Since $X_1, \ldots, X_n$ are i.i.d samples from $\mu$, it yields 
\begin{align*}
\E_{\mu^{\otimes_n}}\Big[M_q^{p/q}\Big(\frac 1n \sum_{i=1}^n \mathcal{N}(\u^\top X_i, \sigma^2)\Big)\Big] &= \frac{2^{q /2}\sigma^q}{\sqrt{\pi}}\Gamma(\frac{q +1}{2})
\E_{\mu^{}}\big[{}_1 F_1\big(-\frac q 2, \frac 12; \frac{-(\u^\top X)^2}{2\sigma^2}\big)\big] \quad (X\sim \mu)\\
&= \frac{2^{q /2}\sigma^q}{\sqrt{\pi}}\Gamma(\frac{q +1}{2}) \sum_{k=0}^\infty \frac{(-\frac q 2)_k}{(\frac 12)_k} \frac{(-1)^k}{(2\sigma^2)^kk!} \E_{\mu^{}}[(\u^\top X)^{2k}]\\
&\leq \frac{2^{q /2}\sigma^q}{\sqrt{\pi}}\Gamma(\frac{q +1}{2}) \sum_{k=0}^\infty \frac{(-\frac q 2)_k}{(\frac 12)_k} \frac{(-1)^k}{(2\sigma^2)^kk!} M_{2k}(\mu).
\end{align*}

Setting $q=2p$ we have $\Delta_n(p, q) = \frac{\log n}{n}$, then 
\begin{align*}
{\bf I} \leq 2^{2p-1}C_{p} \frac{\sigma^{2p}}{\sqrt{\pi}}\Gamma(\frac{2p +1}{2}) \sum_{k=0}^\infty \frac{(-p)_k}{(\frac 12)_k} \frac{(-1)^k}{(2\sigma^2)^kk!} M_{2k}(\mu) \frac{\log(n)}{n}.
\end{align*}

\noindent {\it \underline{Step 2: Control of} ${\bf II}$.}\\

We follow the lines of proofs of Proposition 1 in~\cite{Goldfeld-IEEE20} and Theorem 2 in~\cite{pmlr-v139-nietert21a}. 
Using a coupling $\hat{\hat\mu}_{n}$ and $\mathcal{R}_\u \mu* \mathcal{N}_\sigma)$ via the maximal TV-coupling (see Theorem 6.15 in \cite{villani09optimal}]), the control of the total variation of the Wasserstein distance, we get for any fixed $\u \in \mathbb{S}^{d-1}$
\begin{align*}
\wass^p_p\big( \frac 1n \sum_{i=1}^n \mathcal{N}(\u^\top X_i, \sigma^2), R_\u\mu * \mathcal{N}_\sigma )\big)\leq 2^{p-1} \int_{\R} |t|^p |{h}_{n,\u}(t) - g_\u(t)|dt,
\end{align*}
where $h_{n,\u}$ and $g_\u$ are the densities associated with ${\mu}_{n}$ and $\mathcal{R}_\u \mu* \mathcal{N}_\sigma$, respectively. 
Let $f_{\sigma,\vartheta}$ the probability density function of $\mathcal{N}_{\sigma,\vartheta}$, i.e, $f_{\sigma,\vartheta}(t) = \frac{1}{\sqrt{2\pi(\sigma\vartheta)^2}}e^{-\frac{t^2}{2(\sigma\vartheta)^2}}$ for $\vartheta>0$ to be specified later.
An application of Cauchy-Schwarz inequality gives 
\begin{align*}
\E_{\mu^{\otimes_n}|\mathcal{N}_\sigma^{\otimes_n}}&\Big[\wass^p_p\big( \frac 1n \sum_{i=1}^n \mathcal{N}(\u^\top X_i, \sigma^2), R_\u\mu * \mathcal{N}_\sigma )\big)\Big] \\
&\leq 2^{p-1} \E_{\mu^{\otimes_n}|\mathcal{N}_\sigma^{\otimes_n}} \int_{\R} |t|^p\sqrt{f_{\sigma,\vartheta}(t)} \frac{|h_{n,\u}(t) - g_{\u}(t)|}{\sqrt{f_{\sigma,\vartheta}(t)}}dt\\
&\leq 2^{p-1} \E_{\mu^{\otimes_n}|\mathcal{N}_\sigma^{\otimes_n}} \Big(\int_{\R} |t|^{2p}{f_{\sigma,\vartheta}(t)}dt\Big)^{\frac 12} \Big(\int_{\R}\frac{(h_{n,\u}(t) - g_{\u}(t))^2}{{f_{\sigma,\vartheta}(t)}}dt\Big)^{\frac 12}\\
&\leq 2^{p-1}  \Big(\int_{\R} |t|^{2p}{f_{\sigma,\vartheta}(t)}dt\Big)^{\frac 12}  \Big(\int_{\R}\E_{\mu^{\otimes_n}|\mathcal{N}_\sigma^{\otimes_n}}\frac{(h_{n,\u}(t) - g_\u(t))^2}{{f_{\sigma,\vartheta}(t)}}dt\Big)^{\frac 12}.
\end{align*}
Note that $\int_{\R} |t|^{2p}{f_{\sigma,\vartheta}(t)}dt$ is the $2p$-th moment of $|\mathcal{N}_{\sigma,\vartheta}(t)$ equals to (see Equation (18) in~\cite{winkelbauer2014moments}) 
\begin{align*}
\int_{\R} |t|^{2p}{f_{\sigma,\vartheta}(t)}dt =  \frac{(\sigma\vartheta)^{2p} 2^p}{\sqrt{\pi}} \Gamma\big(\frac{2p+1}{2}\big).
 \end{align*} 
Moreover,
\begin{align*}
h_{n,\u}(t)= \frac 1n \sum_{i=1}^n \diff \mathcal{N}(\u^\top X_i, \sigma^2)(t) = \frac 1n \sum_{i=1}^n f_{\sigma,\vartheta}(t - \u^\top X_i),
\end{align*}
It is clear to see that  $h_{n,\u}(t)$ is a sum of i.i.d. terms with expectation $g_\u(t)$, which implies 
\begin{align*}
\E_{\mu^{\otimes_n}|\mathcal{N}_\sigma^{\otimes_n}}\big[(h_{n,\u}(t) - g_{\u}(t))^2\big] &= \V_{\mu^{\otimes_n}}\Big[\frac 1n \sum_{i=1}^n f_{\sigma,\vartheta}(t - \u^\top X_i)\Big]\\
& = \frac 1n \V_{\mu^{}}[f_{\sigma,\vartheta}(t - \u^\top X]\\
& \leq \frac 1n \E_{\mu^{}}[(f_{\sigma,\vartheta}(t - \u^\top X)^2]\\
& \leq \frac{(2\pi\sigma^2)^{-1}}{n} \E_{\mu^{}}[e^{\frac{-1}{\sigma^2}(t - \u^\top X)^2}].
\end{align*}
Now 
\begin{align*}
\E_{\mu^{}}[e^{\frac{-(t - \u^\top X)^2}{\sigma^2}}]  &= \int_{\norm{x} \leq \frac{|t|}{2}} e^{\frac{-1}{\sigma^2}(t - \u^\top x)^2}\diff\mu(x) + \int_{\norm{x} > \frac{|t|}{2}} e^{\frac{-1}{\sigma^2}(t - \u^\top x)^2}\diff\mu(x).
\end{align*}
Remark that when $\norm{x} \leq \frac{|t|}{2}$, then $(t - \u^\top X)^2 \geq |t|^2 - |\u^\top x|^2 \geq |t|^2 - \norm{x}^2$ (since $\norm{u}^2 = 1$). We get $(t - \u^\top X)^2 \geq \frac{|t|^2}{4}$ and hence 
\begin{align*}
\int_{\norm{x} \leq \frac{|t|}{2}} e^{\frac{-1}{\sigma^2}(t - \u^\top x)^2}\diff\mu(x) \leq e^{\frac{-t^2}{4\sigma^2}} \text{ and } \int_{\norm{x} > \frac{|t|}{2}} e^{\frac{-1}{\sigma^2}(t - \u^\top x)^2}\diff\mu(x) \leq \P\big[\norm{X} > \frac{|t|}{2}\big]
\end{align*}

This gives,
\begin{align*}
\int_{\R}\E_{\mu^{\otimes_n}|\mathcal{N}_\sigma^{\otimes_n}}\frac{(h_{n,\u}(t) - g_\u(t))^2}{{f_{\sigma,\vartheta}(t)}}\diff t \leq \frac{(2\pi\sigma^2)^{-1}(\sqrt{2\pi}\sigma\vartheta)}{n}\Big( \int_{\R}e^{\frac{t^2}{2(\sigma\vartheta)^2}}e^{\frac{-t^2}{4\sigma^2}}\diff t + \int_{\R}e^{\frac{t^2}{2(\sigma\vartheta)^2}}\P\big[\norm{X} > \frac{|t|}{2}\big]\diff t\Big).
\end{align*}
Note that the integral $\int_{\R}e^{\frac{t^2}{2(\sigma\vartheta)^2}}e^{\frac{-t^2}{4\sigma^2}}\diff t = \int_{\R}e^{-\big(\frac12 - \frac{1}{\vartheta^2}\big)\frac{t^2}{2\sigma^2}}\diff t$ is finite if and only if  $\frac12 - \frac{1}{\vartheta^2} > 0$ namely $\vartheta > \sqrt{2}$ and its value is given by 
\begin{align*}
\int_{\R}e^{\frac{t^2}{2(\sigma\vartheta)^2}}e^{\frac{-t^2}{4\sigma^2}}\diff t = \sqrt{\frac{2\pi\sigma^2}{\frac 12 - \frac{1}{\vartheta^2}}} = \sqrt{\frac{4 \pi \sigma^2\vartheta^2}{\vartheta^2 - 2}}.
\end{align*}
For the second integral 
\begin{align*}
\int_{\R}e^{\frac{t^2}{2(\sigma\vartheta)^2}}\P\big[\norm{X} > \frac{|t|}{2}\big]\diff t = 2 \int_{0}^\infty e^{\frac{t^2}{2(\sigma\vartheta)^2}}\P\big[\norm{X} > \frac{t}{2}\big]\diff t = 4 \int_{0}^\infty e^{\frac{2\xi^2}{\sigma^2\vartheta^2}}\P\big[\norm{X} > \xi\big]\diff \xi
\end{align*}
Then,
\begin{align*}
{\bf II} 
&\leq n^{-1/2} 4^{p-1} \Big\{(2\pi\sigma^2)^{-1}(\sqrt{2\pi}\sigma\vartheta)\frac{(\sigma\vartheta)^{2p} 2^p}{\sqrt{\pi}} \Gamma\big(\frac{2p+1}{2}\big) \Big\}^{\frac 12}\Big(\sqrt{\frac{4 \pi \sigma^2\vartheta^2}{\vartheta^2 - 2}} + 4 \int_{0}^\infty e^{\frac{2\xi^2}{\sigma^2\vartheta^2}}\P\big[\norm{X} > \xi\big]\diff \xi\Big)^{\frac 12}.
\end{align*}

this gives the desired result.

\subsubsection{Proof of Proposition~\ref{theorem:first_control_sample_complexity_for_mu_and_nu}}

Using triangle inequality, we have 
\begin{align*}
{\wass_p}(\hat{\hat\mu}_{n},\hat{\hat\nu}_{n}) \leq {\wass_p}(\hat{\hat\mu}_{n}, \mathcal{R}_\u \mu* \mathcal{N}_\sigma) + {\wass_p}(\mathcal{R}_\u \mu* \mathcal{N}_\sigma, \mathcal{R}_\u \nu* \mathcal{N}_\sigma) + {\wass_p}(\mathcal{R}_\u \nu* \mathcal{N}_\sigma, \hat{\hat\nu}_{n}). 
\end{align*}
and then
\begin{align*}
\wass_p^p(\hat{\hat\mu}_{n},\hat{\hat\nu}_{n}) \leq 3^{p-1} \big\{\wass_p^p(\hat{\hat\mu}_{n}, \mathcal{R}_\u \mu* \mathcal{N}_\sigma) + \wass_p^p(\mathcal{R}_\u \mu* \mathcal{N}_\sigma, \mathcal{R}_\u \nu* \mathcal{N}_\sigma) + \wass_p^p(\mathcal{R}_\u \nu* \mathcal{N}_\sigma, \hat{\hat\nu}_{n})\big\}. 
\end{align*}
This implies that 
\begin{align*}
\E_{\mu^{\otimes_n}| \mathcal{N}_\sigma^{\otimes_n}}
&\E_{\nu^{\otimes_n}| \mathcal{N}_\sigma^{\otimes_n}}[{\hatgssw^p}({\hat\mu}_n,\hat\nu_n)]\\
& \leq 3^{p-1} {\gssw^p}({\mu},\nu)  +  3^{p-1}\E_{\mu^{\otimes_n}| \mathcal{N}_\sigma^{\otimes_n}}[{\hatgssw^p}({\hat\mu}_n,\mu)] + 3^{p-1}\E_{\nu^{\otimes_n}| \mathcal{N}_\sigma^{\otimes_n}}[{\hatgssw^p}({\hat\nu}_n,\nu)].
\end{align*}
From Proposition~\ref{theorem:sample_double_empirical_complexity}, we have 
\begin{align*}
\E_{\mu^{\otimes_n}|\mathcal{N}_\sigma^{\otimes_n}}[{\hatgssw^p}({\hat\mu}_n,\mu)] \leq \Xi_{p,\sigma, \vartheta} \frac{1}{\sqrt{n}}+ \Upsilon_{p,\sigma,\vartheta, \mu} \frac{\log n}{n}.
\end{align*}
Similarly,
\begin{align*}
\E_{\nu^{\otimes_n}|\mathcal{N}_\sigma^{\otimes_n}}[{\hatgssw^p}({\hat\nu}_n,\nu)] \leq \Xi_{p,\sigma, \vartheta} \frac{1}{\sqrt{n}}+ \Upsilon_{p,\sigma,\vartheta, \nu} \frac{\log n}{n}.
\end{align*}
This gives that 
\begin{align*}
\E_{\mu^{\otimes_n}| \mathcal{N}_\sigma^{\otimes_n}}
\E_{\nu^{\otimes_n}| \mathcal{N}_\sigma^{\otimes_n}}[{\hatgssw^p}({\hat\mu}_n,\hat\nu_n)]\leq 3^{p-1} {\gssw^p}({\mu},\nu) +  3^{p}\Xi_{p,\sigma, \vartheta} \frac{1}{\sqrt{n}} + 3^{p-1}(\Upsilon_{p,\sigma,\vartheta, \mu} + \Upsilon_{p,\sigma,\vartheta, \nu}) \frac{\log n}{n}
\end{align*}
This ends the proof of the first statement in~Proposition~\ref{theorem:first_control_sample_complexity_for_mu_and_nu}.
For the second one, we also use a triangle inequality
\begin{align*}
\wass_p^p(\mathcal{R}_\u \mu* \mathcal{N}_\sigma, \mathcal{R}_\u \nu* \mathcal{N}_\sigma)
\leq 3^{p-1} \big\{\wass_p^p(\mathcal{R}_\u \mu* \mathcal{N}_\sigma,\hat{\hat\mu}_{n})+ \wass_p^p(\hat{\hat\mu}_{n},\hat{\hat\nu}_{n}) + \wass_p^p(\hat{\hat\nu}_{n}),\mathcal{R}_\u \nu* \mathcal{N}_\sigma\big\}.
\end{align*}
Then we control each term as we did before.

\subsection{Proof of Proposition~\ref{theorem:error_MC}: projection complexity} \label{sec:proof_of_theorem_theorem:error_mc}

Using Holder's inequality, we have 
\begin{align*}
\E_{\u \sim u_d}\big[\big|\widehat{\gssd^p}(\mu,\nu) - {\gssd^p}(\mu,\nu)\big|\big]
&\leq \Big(\E_{\u \sim u_d}\big[\big|\widehat{\gssd^p}(\mu,\nu) - {\gssd^p}(\mu,\nu)\big|^2\big]\Big)^{1/2}\\
&= \Big(\V_{\u \sim u_d}\big[\big|\widehat{\gssd^p}(\mu,\nu)\big|\big]\Big)^{1/2}\\
&= \Big(\V_{\u \sim u_d}\big[\big|{\gssd^p}(\mu,\nu)\big|\big]\Big)^{1/2}\\
&= \frac{A(p,\sigma)}{\sqrt{L}}.
\end{align*}

\subsection{Proof of Proposition~\ref{proposition:2-level-noise}}

For all $\u\in\mathbb{S}^{d-1}$ we have $\mathcal{R}_\u \mu, \mathcal{R}_\u \nu \in \mathcal{P}(\R)$. By application of the inequality of noise level satisfied by $\divg$ in one dimension we get 
\begin{equation*}
\divg^p( \mathcal{R}_\u \mu * \mathcal{N}_{\sigma_2}, \mathcal{R}_\u \nu * \mathcal{N}_{\sigma_2}) \leq 
\divg^p(\mathcal{R}_\u \mu * \mathcal{N}_{\sigma_1}, \mathcal{R}_\u \nu * \mathcal{N}_{\sigma_1}).   
\end{equation*}
Then, computing the expectation over the projections $\u$ since the divergence is non-negative 
concludes the proof.

 \subsection{Proof of Proposition~\ref{proposition:GS-SWD_sigma_1_2}: relation between $\gssw^p(\mu,\nu)$ under two noise levels} \label{sub:proof_of_proposition_proposition:gs-swd_sigma_1_2}

The proof follows the same lines in proof of Lemma 1 in~\cite{pmlr-v139-nietert21a}.
First, we have that $\mathcal{N}_{\sigma_2} = \mathcal{N}_{\sigma_1} * \mathcal{N}_{\sqrt{\sigma_2^2 - \sigma_1^2}}$. Setting the following random variables: $X_\u\sim \mathcal{R}_\u \mu, Y_\u\sim\mathcal{R}_\u \nu, Z_X \sim \mathcal{N}_{\sigma_1}, Z_Y \sim \mathcal{N}_{\sigma_1}, Z'_X \sim \mathcal{N}_{\sqrt{\sigma_2^2 - \sigma_1^2}}, Z'_Y \sim \mathcal{N}_{\sqrt{\sigma_2^2 - \sigma_1^2}}$. The sliced Wasserstein distance $\text{W}_p^p(\mathcal{R}_\u \mu * \mathcal{N}_{\sigma_2},\mathcal{R}_\u \nu* \mathcal{N}_{\sigma_2})$ is given as a minimization over couplings $(X_\u, Z_X, Z'_X)$ and $(Y_\u, Z_Y, Z'_Y)$. Using the inequality $\E[|X|^p] - 2^{p-1}\E[|Y|^p]\leq 2^{p-1}\E[|X+Y|^p]$ for any random variables $X, Y \in \mathbb{L}_p$ integrable, we obtain, 
\begin{align*}
2^{p-1}\E\big[|(X_\u + Z_X) - (Y_\u + Z_Y) + (Z'_X + Z'_Y)|^p\big]
&\geq \E\big[|(X_\u + Z_X) - (Y_\u + Z_Y) |^p\big] -2^{p-1}\E\big[|(Z'_X + Z'_Y)|^p\big]\big).
\end{align*}
Hence,
\begin{align*}
2^{p-1}\text{W}^p_p(\mathcal{R}_\u \mu * \mathcal{N}_{\sigma_2},\mathcal{R}_\u \nu* \mathcal{N}_{\sigma_2})
&\geq \inf\Big(\E\big[|(X_\u + Z_X) - (Y_\u + Z_Y) |^p\big] -2^{p-1}\E\big[|(Z'_X + Z'_Y)|^p\big]\big)\Big)\\
&\geq \text{W}^p_p(\mathcal{R}_\u \mu * \mathcal{N}_{\sigma_1},\mathcal{R}_\u \nu* \mathcal{N}_{\sigma_1}) - 2^{p-1}\sup\E\big[|(Z'_X + Z'_Y)|^p\big]\\
&\geq \text{W}^p_p(\mathcal{R}_\u \mu * \mathcal{N}_{\sigma_1},\mathcal{R}_\u \nu* \mathcal{N}_{\sigma_1}) - 2^{2p-1} \sup\E\big[|(Z'_X)|^p\big].
\end{align*}
Therefore,
\begin{align*}
2^{p-1} \gsswww^p(\mu,\nu) &\geq \gssww^p(\mu,\nu) - 2^{2p-1} \sup\E\big[|(Z'_X)|^p\big].
\end{align*}
Hence,
\begin{align*}
\gssww^p(\mu,\nu) &\leq 2^{p-1} \gsswww^p(\mu,\nu) + 2^{2p-1} \sup\E\big[|(Z'_X)|^p\big]
\end{align*}
then
\begin{align*}
	\gssww^p(\mu,\nu) &\leq 2^{p-1} \gsswww^p(\mu,\nu)+ 2^{\frac{5p}{2}} (\sigma_2^2 - \sigma_1^2)^{p},
\end{align*}
and concludes the proof.

\subsection{Proof of Proposition~\ref{prop:continuity_of_gs_swd}: continuity of the  smoothed Gaussian sliced Wasserstein w.r.t. $\sigma$}

From Lemma 1 in~\citep{pmlr-v139-nietert21a}, we know that the Gaussian-smoothed Wasserstein is continuous with respect to $\sigma$, for any distribution $\mathcal{R}_\u\nu$ and $\mathcal{R}_\u \mu$. In addition, for any $\u$,
	we have $\wass_p(\mathcal{R}_\u\nu * \mathcal{N}_\sigma,
	\mathcal{R}_\u\mu * \mathcal{N}_\sigma) \leq 
	\wass_p(\mathcal{R}_\u\nu ,
	\mathcal{R}_\u\mu)$. Then by applying Lebesgue's dominated convergence theorem~\citep{bowers2014introductory} to the above inequality with $\wass_p(\mathcal{R}_\u\nu ,
	\mathcal{R}_\u\mu)$ as a dominating function, that is $u_d$-almost everywhere integrable because both measures are in $\mathcal{P}_p(\R^d)$, we then conclude that the Gaussian-smoothed SWD is continuous w.r.t. $\sigma$.

\subsection{Proof of Proposition~\ref{prop:continuity_smoothed_mmd}: continuity of the  smoothed sliced squared-MMD w.r.t. $\sigma$}
\label{sub:proof_of_proposition_prop:continuity_smoothed_mmd}
Let us first recall the definition of the MMD divergence. 
Let $k: \R\times \R\rightarrow \R$ be a measurable bounded kernel on $\R$ and consider the reproducing kernel Hilbert space (RKHS) $\mathcal{H}_k$ associated with $k$ and equipped with inner product $<\cdot, \cdot>_{\mathcal{H}_k}$ and norm $\norm{\cdot}_{\mathcal{H}_k}$. Let $\mathcal{P}_{\mathcal{H}_k}(\R)$ be the set of probability measures $\eta$ such that $\int_\R\sqrt{k(t,t)} d\eta(x) < \infty.$ The kernel mean embedding is defined as $\Phi_k(\eta) = \int_{\R} k(\cdot, t)d\eta(t).$ The squared-maximum mean discrepancy between $\eta, \zeta\in \mathcal{P}(\R)$ denoted as $\MMD:\mathcal{P}_{\mathcal{H}_k}(\R)\times \mathcal{P}_{\mathcal{H}_k}(\R) \rightarrow \R_+$ is  expressed as the distance between two such kernel mean embeddings. It is defined as~\cite{gretton2012kernel}
\begin{align*}
\MMD^2(\eta,\zeta)&= \left \|\Phi_k(\eta) - \Phi_k(\zeta)\right \|_{\mathcal{H}_k}^2 = \E_{T, T'\sim \eta}[k(T,T')] - 2 \E_{T\sim\eta, R\sim\zeta}[k(T,R)] + \E_{R, R'\sim \zeta}[k(R,R')]
\end{align*}
where $T$ and $T'$ are independent random variables drawn according to $\eta$, $R$ and $R'$ are independent random variables
drawn according to $\zeta$, and $T$ is independent of $R$. We define the Gaussian Smoothed Sliced squared-$\MMD$ as follows:
\begin{align*}
\gssmmd^2(\mu,\nu) &= \int_{\mathbb{S}^{d-1}}\left \|\Phi_k(\mathcal{R}_\u \mu * \mathcal{N}_\sigma) - \Phi_k(\mathcal{R}_\u \nu * \mathcal{N}_\sigma)\right \|_{\mathcal{H}_k}^2u_d(\u)\diff\u\nonumber\\ &= \int_{\mathbb{S}^{d-1}}\big(\E_{T, T'\sim \mathcal{R}_\u \mu * \mathcal{N}_\sigma}[k(T,T')] - 2 \E_{T\sim\mathcal{R}_\u \mu * \mathcal{N}_\sigma, R\sim\mathcal{R}_\u \nu * \mathcal{N}_\sigma}[k(T,R)]\nonumber\\
&\hspace{2cm} + \E_{R, R'\sim \mathcal{R}_\u \nu * \mathcal{N}_\sigma}[k(R,R')]\big)u_d(\u)\diff\u.
\end{align*}

From the definition of the smoothed sliced squared-MMD, we have \begin{align*}
\E_{T, T'\sim \mathcal{R}_\u \mu * \mathcal{N}_\sigma}[k(T,T')] &= \iint_{\R\times\R}k(t,t')\diff\mathcal{R}_\u \mu * \mathcal{N}_\sigma(t)\diff\mathcal{R}_\u \mu * \mathcal{N}_\sigma(t')\\
&= \iint_{\R\times\R} \Big(\int_\R k(t+z,t')\diff\mathcal{R}_\u \mu(z)\mathcal{N}_\sigma(t)\Big)\diff\mathcal{R}_\u \mu * \mathcal{N}_\sigma(t')\\
&= \iint_{\R\times\R} \Big(\int_{\R^d} k(t+\u^\top x,t')\diff\mu(x)\mathcal{N}_\sigma(t)\Big)\diff\mathcal{R}_\u \mu * \mathcal{N}_\sigma(t')\\
&= \iint_{\R\times\R} \iint_{\R^d\times \R^d} k(t+\u^\top x,t'+\u^\top x')\diff\mu(x)\diff\mu(x')\diff\mathcal{N}_\sigma(t) \diff\mathcal{N}_\sigma(t').
\end{align*}
Similarly, 
\begin{align*}
\E_{R, R'\sim \mathcal{R}_\u \nu * \mathcal{N}_\sigma}[k(R,R')] = \iint_{\R\times\R} \iint_{\R^d\times \R^d} k(r+\u^\top y,r'+\u^\top y')\diff\nu(y)\diff\nu(y')\diff\mathcal{N}_\sigma(r) \diff\mathcal{N}_\sigma(r')
\end{align*}
and
\begin{align*}
\E_{T\sim\mathcal{R}_\u \mu * \mathcal{N}_\sigma, R\sim\mathcal{R}_\u \nu * \mathcal{N}_\sigma}[k(T,R)]  = \iint_{\R\times\R} \iint_{\R^d\times \R^d} k(t+\u^\top x,r+\u^\top y)\diff\mu(x)\diff\nu(y)\diff\mathcal{N}_\sigma(t) \diff\mathcal{N}_\sigma(r).
\end{align*}
Together the assumption of boundness of the kernel function $k$ and the continuity of integrals, 
the three latter terms are continuous functions w.r.t. $\sigma \in (0, \infty).$
Again by the boundness of the kernel function $k$, there exists a positive finite constant $C_k$ such that 
\begin{equation*}
\big|\E_{T, T'\sim \mathcal{R}_\u \mu * \mathcal{N}_\sigma}[k(T,T')] -2\E_{T\sim\mathcal{R}_\u \mu * \mathcal{N}_\sigma, R\sim\mathcal{R}_\u \nu * \mathcal{N}_\sigma}[k(T,R)] + \E_{R, R'\sim \mathcal{R}_\u \nu * \mathcal{N}_\sigma}[k(R,R')]\big|\leq 4C_k. \end{equation*}
We conclude the continuity of $\sigma\mapsto \gssmmd^2(\mu,\nu)$ by an application of the continuity of integrals.

\section{Additional experiments} \label{sec:additional_experiments}

\subsection{Sample complexity on CIFAR dataset}
We have also evaluated the sample complexity for the CIFAR dataset by sampling sets of increasing size.
Results reported in Figure \ref{fig:cifar} confirms the findings obtained from the toy dataset. 
\begin{figure}[h]
	\centering
		\includegraphics[width=5cm]{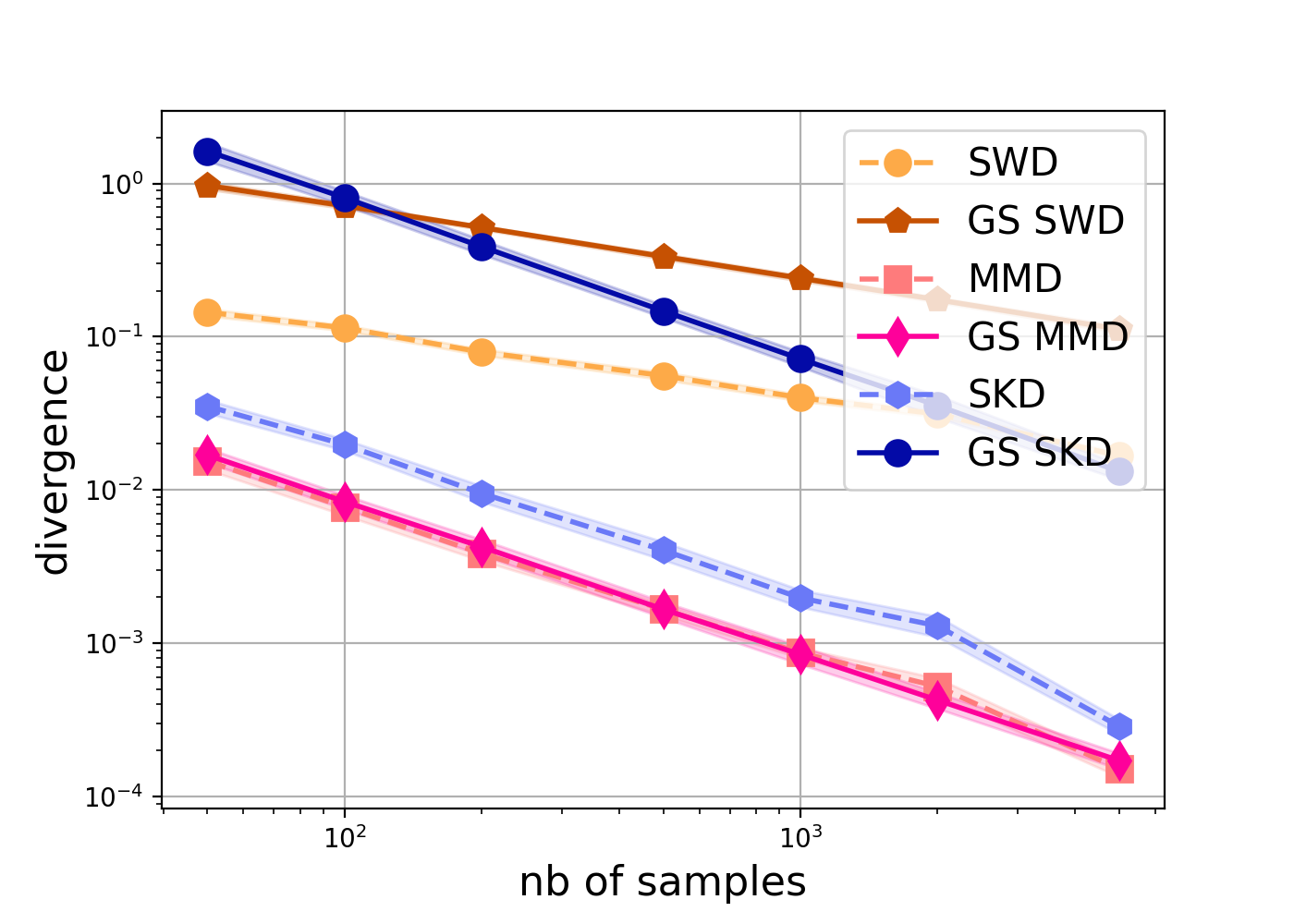} 
	\caption{Measuring the divergence between two sets of samples drawn iid from the CIFAR10
		dataset.
		We compare three sliced divergences and their Gaussian smoothed versions
		with a $\sigma=3$.\label{fig:cifar}}
\end{figure}

\subsection{Identity of indiscernibles}
The second experiment aims at checking whether our divergences converge towards a small value
when the distributions to be compared are the same. For this, { we consider samples from distributions $\mu$ and $\nu$  chosen as normal distributions with respectively}   mean $2 \times \mathbf{1}_d$ and $s \mathbf{1}_d$ with
varying $s$ (noted as the displacement). Results are depicted in Figure \ref{fig:divdisplacement}. We can see
that all methods are able to attain their minimum when $s=2$. Interestingly, the gap between
the Gaussian smoothed and non-smoothed divergences for Wasserstein and Sinkhorn is almost
indiscernible as the distance between distribution increases.

\begin{figure}[H]
	\centering
	\includegraphics[width=5cm]{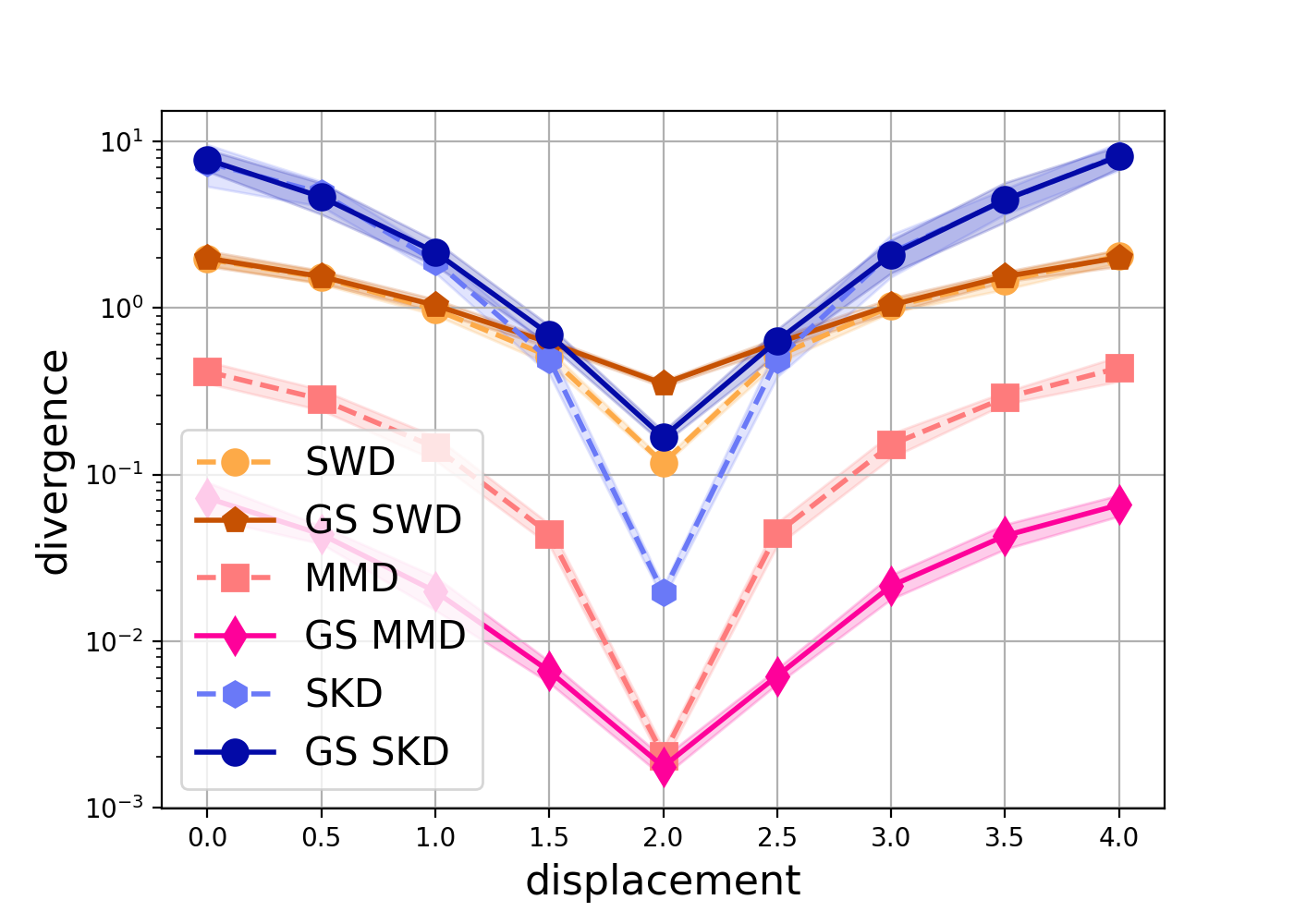}   
	\caption{Measuring the divergence between two sets of samples in $\R^{50}$, one
		with mean {$2  \mathbf{1}_d$} and the other with mean {$s \mathbf{1}_d$} 		with increasing $s$.
		We compare three sliced divergences and their Gaussian smoothed version
		with a $\sigma=3$.
		\label{fig:divdisplacement}}  
\end{figure}

\clearpage
\bibliography{refs}
\bibliographystyle{tmlr} 
\clearpage

\end{document}